\documentclass[11pt]{article}

\usepackage[preprint]{acl}

\usepackage{times}
\usepackage{latexsym}
\usepackage{booktabs}     
\usepackage{array}   
\usepackage{tabularx}
\usepackage{hyperref}
\usepackage{fontawesome}
\newcommand{\ExternalLink}{\faExternalLink}
\usepackage{svg}
\usepackage{amsmath}
\usepackage{xcolor}
\usepackage{float}
\usepackage{varwidth}

\usepackage[T1]{fontenc}

\usepackage[utf8]{inputenc}

\usepackage{microtype}

\usepackage{inconsolata}

\usepackage{graphicx}

\title{SwissGov-RSD:\\A Human-annotated, Cross-lingual Benchmark for Token-level\\\textit{R}ecognition of \textit{S}emantic \textit{D}ifferences Between Related Documents}

\author{Michelle Wastl \quad Jannis Vamvas \quad Rico Sennrich
\vspace{0.1cm}\\
 Department of Computational Linguistics, University of Zurich\\
\texttt{\{michelle.wastl,jannisnikos.vamvas,rico.sennrich\}@uzh.ch}
}

\begin{document}
\maketitle
\begin{abstract}
Recognizing semantic differences across documents is crucial for text generation evaluation and content alignment, especially in cross-lingual settings. However, as a standalone task, it has received little attention. We address this by introducing \textit{SwissGov-RSD}, the first naturalistic, document-level, cross-lingual dataset for semantic difference recognition. It encompasses a total of 224 multi-parallel documents in English--German, English--French, and English--Italian with token-level difference annotations by human annotators.
We evaluate a variety of open-source and closed-source large language models as well as encoder models across different fine-tuning settings on this new benchmark. 
Our results show that current automatic approaches perform poorly compared to their performance on monolingual, sentence-level, and synthetic benchmarks, revealing a considerable gap for both LLMs and encoder models. 
We make our code and dataset publicly available\footnote{Code: \href{https://github.com/ZurichNLP/SwissGov-RSD}{https://github.com/ZurichNLP/SwissGov-RSD}; \\ Data: \href{https://huggingface.co/datasets/ZurichNLP/SwissGov-RSD}{https://huggingface.co/datasets/ZurichNLP/SwissGov-RSD}. We release the SwissGov-RSD labels under a CC BY license, while the copyright of the text remains with the Swiss federal authorities:~\href{https://www.admin.ch/gov/en/start/terms-and-conditions.html}{copyright notice}.}.
\end{abstract}

\section{Introduction}

\begin{figure}[t!]
  \centering
  \includegraphics[width=\linewidth]{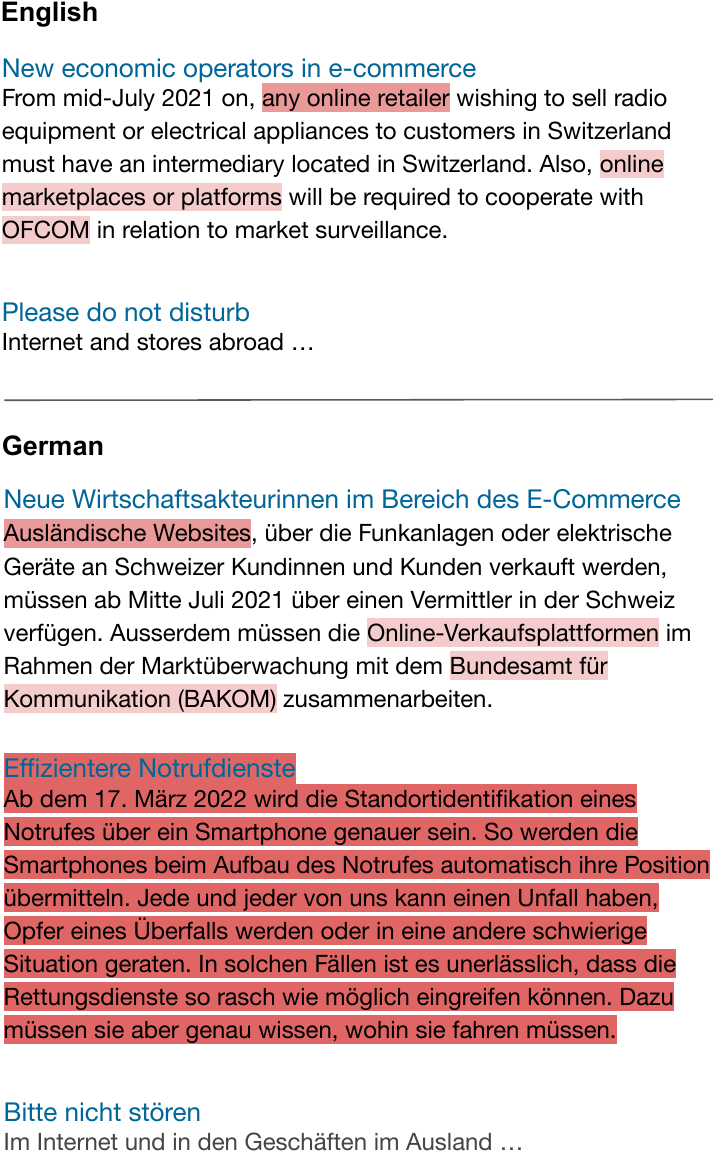}
  \caption{Excerpt from an English--German document pair with difference annotations from the SwissGov-RSD dataset. The differences range from small semantic shifts and explicitations to omitted paragraphs. The saturation of the red highlight indicates how large the semantic difference is, i.e., the paragraph marked in deep red contains information about emergency calls and is completely omitted in the English document.}
  \label{fig:admin-example}
\end{figure}

Recognizing semantic differences (RSD) across different versions of a text---whether monolingual or cross-lingual---is an underexplored challenge in natural language understanding. Beyond theoretical interest in tasks such as machine translation evaluation~\citep{burchardt-2013-multidimensional, freitag-etal-2021-experts, rei-etal-2023-inside} and interpretable semantic similarity (iSTS)~\citep{agirre-etal-2016-semeval, agirre-etal-2016-semeval-2016}, it has high practical relevance. For example, multilingual websites, especially in public and governmental contexts, are expected to convey equivalent content across languages. Yet in practice, discrepancies can go unnoticed, as illustrated in Figure~\ref{fig:admin-example} with excerpts from the multilingual Swiss government portal \href{https://www.admin.ch}{admin.ch}.

\begin{table*}[h]
\centering
\small
\begin{tabular}{@{}lrrrrr@{}}
\toprule
\textbf{Dataset} & \textbf{\# Doc Pairs} & \textbf{\# Total Tokens} & \textbf{Min. Doc Length} & \textbf{Max. Doc Length} & \textbf{Avg. Doc Length} \\
\midrule
SwissGov-RSD EN-DE & 224 & 173,043 & 50 & 2,528 & 386 \\
SwissGov-RSD EN-FR & 224 & 195,930 & 52 & 2,837 & 437 \\
SwissGov-RSD EN-IT & 224 & 187,115 & 60 & 2,577 & 418 \\
\bottomrule
\end{tabular}
\caption{General statistics for SwissGov-RSD. Length is measured in tokens separated by whitespaces.}
\label{tab:data-length}
\end{table*}

We collect document pairs with naturally occurring semantic differences from this portal across its language versions and release them with human-annotated token-level difference spans in a new multi-parallel dataset: \textit{SwissGov-RSD}. We then evaluate a range of systems, spanning unsupervised, few-shot and fine-tuned settings, on their ability to automatically detect these differences. Finally, we compare system performance on our dataset to results on the synthetically constructed iSTS-RSD benchmark~\citep{vamvas-sennrich-2023-towards} and explore the limitations of current systems.

Our contributions are the following:
\begin{itemize}
    \item We construct and release SwissGov-RSD, the first human-annotated, document-level, cross-lingual dataset designed for token-level difference recognition for three language pairs: English--German, English--French, and English--Italian.
    \item We benchmark 6 LLMs and 12 encoder models using 3 different approaches (Figure~\ref{fig:schem-example}) in combination with differing degrees of supervision on the naturalistic SwissGov-RSD and the synthetically constructed iSTS-RSD.
    \item We reveal a considerable gap between the performance on the naturalistic and synthetic datasets, showing that all systems perform substantially worse when evaluated on the naturalistic data with differences in Spearman correlation of up to 78.
\end{itemize}

\noindent{}Our results highlight the need for more specialized approaches to make semantic difference recognition applicable in real-world settings and point towards the limitations of synthetic benchmarking in general, where strong performance can mask a failure to generalize.

\section{Recognition of Semantic Differences}

Recognizing Semantic Differences (RSD) concerns identifying which parts of two texts differ in meaning. Rather than assigning a single similarity score to a text pair, RSD targets fine-grained, token-level differences. This framing is particularly relevant for comparing document versions, either across time or languages, where differences may range from minor reformulations to omitted paragraphs, as illustrated in Figure~\ref{fig:admin-example}.

To formalize the task, \citet{vamvas-sennrich-2023-towards} propose treating RSD as a token-level regression problem, where each token is assigned a score indicating its semantic divergence from its aligned counterpart. Their benchmark repurposes human span-level similarity annotations from iSTS~\citep{agirre-etal-2016-semeval}. They synthetically augment the iSTS dataset through extension with \mbox{PAWS(-X)}~\citep{zhang-etal-2019-paws, yang-etal-2019-paws}, concatenation, reordering and machine translation to encompass cross-linguality and emulate document structures. In the remainder of this paper, we will refer to this benchmark as \textit{iSTS-RSD}.

Initial evaluations on iSTS-RSD by~\citet{vamvas-sennrich-2023-towards} suggest that while longer texts increase the task difficulty, cross-lingual scenarios are the most challenging, highlighting the need for broader evaluation settings and more realistic data to evaluate on.

In this work, we address this need by assembling a human-annotated dataset inspired by realistic use cases and therefore enabling evaluation in settings that reflect real-world, cross-lingual applications.

\section{SwissGov-RSD}



Switzerland publishes its government websites at \href{https://www.admin.ch/gov/de/start.html}{admin.ch} in English and the country's four official languages: German, French, Italian, and Romansh. These websites provide coherent texts that span a wide range of topics, making them a valuable source of natural multi-parallel data, which, either due to translation errors or unsynchronized content updates, contain semantic differences across language versions. 
We therefore collect multi-parallel documents from \href{https://www.admin.ch/gov/de/start.html}{admin.ch} and its subdomains in English, German, French, and Italian\footnote{We exclude Romansh due to its limited availability, which would have substantially reduced dataset size.}. The resulting dataset includes aligned document pairs in EN--DE, EN--FR, and EN--IT with descriptive statistics shown in Table~\ref{tab:data-length}. 

\begin{table}[t]
\centering
\resizebox{\columnwidth}{!}{
\begin{tabular}{@{}lrrr@{}}
\toprule
\textbf{Metric} & \textbf{EN--DE} & \textbf{EN--FR} & \textbf{EN--IT} \\
\midrule
Total \# spans A1 & 348 & 148 & 276 \\
Total \# spans A2 & 139 & 192 & 382 \\
Total \# differences in tokens A1 & 2,543 & 1,167 & 4,451 \\
Total \# differences in tokens A2 & 1,699 & 1,453 & 3,420 \\
\midrule
\# fuzzy matched span pairs >= 50 & 145 & 54 & 222 \\
\# fuzzy matched span pairs >= 75 & 52 & 36 & 85 \\
\# fuzzy matched span pairs >= 90 & 37 & 28 & 73 \\
\# exactly matching span pairs & 27 & 23 & 64 \\
\midrule
Mean IoU (English) & 23.98 & 24.06 & 32.10 \\
Mean IoU (other language) & 39.95 & 29.65 & 43.52 \\
Mean F1 (English) & 34.61 & 33.34 & 44.15 \\
Mean F1 (other language)& 51.14 & 40.04 & 55.60 \\
\midrule
Corr. fuzzy matched span pairs >= 50 & 76.47 & 90.21 & 88.12 \\
Corr. fuzzy matched span pairs >= 75 & 78.36 & 98.05 & 93.71 \\
Corr. fuzzy matched span pairs >= 90 & 79.50 & 97.66 & 92.89 \\
Corr. exactly matched span pairs & 73.27 & 91.09 & 82.76 \\
\midrule
Token-level Spearman correlation & 56.14 & 43.61 & 65.09 \\
\bottomrule
\end{tabular}
}
\caption{Statistics and agreement between annotators in 25 overlapping documents.}
\label{tab:main-round}
\end{table}

\paragraph{Scraping and Preprocessing} We crawl \href{admin.ch}{admin.ch} and its subdomains, which reflect different topics\footnote{See Appendix~\ref{app:admin-domains} for an overview of the included subdomains.}, in English. If a given page includes links to corresponding versions in German, French, and Italian, we consider it for our dataset. We extract the HTML content for all four languages and apply the following filters: documents must contain at least three sentences of natural language text in the correct language; pages that are non-linguistic (e.g., link-only pages or templates) are discarded. After filtering, we retain 235 multi-parallel documents.

\vspace{0.3cm}
\noindent\textbf{Labeling Scheme} Following iSTS(-RSD)~\citep{agirre-etal-2016-semeval-2016, vamvas-sennrich-2023-towards}, we annotate semantic differences at the span level. Each annotation consists of at least one token in one of the texts, optionally aligned with a span in the other language. Spans need not be identical in length, position, or continuity across languages. For labeling, we adopt the inverse of the five-point scale previously used by~\citet{agirre-etal-2016-semeval, agirre-etal-2016-semeval-2016}, due to our focus on semantic differences rather than explaining semantic similarity estimates, such that label 1 indicates minimal difference (minor unimportant detail), and 5 denotes complete semantic dissimilarity. Unannotated text is assumed to be semantically equivalent (label = 0). The scale is designed to support fine-grained cross-lingual comparison and accommodate both symmetric differences consisting of alignable token spans in both texts and asymmetric ones, where text from one side has no correspondence on the other side (e.g., omission/addition). 

\paragraph{Annotation Process} To annotate the collected data, we recruit two undergraduate/bachelor's students in computational linguistics for each language pair. The annotators are either native speakers or highly proficient in the languages they are assigned to. 
All annotators receive detailed written guidelines outlining the annotation procedure and the semantic difference labeling scheme\footnote{ Appendix~\ref{app:guidelines} contains the full guidelines, including a screenshot of the interface and examples to illustrate the labeling scheme.}.

The annotation is conducted in two phases: a trial phase and a main phase. During the trial phase, all annotators work on the same three text pairs for their assigned language pair. These annotations are evaluated both automatically (Table~\ref{tab:trial_round}) and manually, and targeted feedback is provided based on the results.
In the main phase, each annotator is assigned half the text pairs in a language pair with an additional 25 samples that overlap with the other annotator. The shared samples are used for the annotation validation described below. 

An interactive annotation interface specifically developed for this task is used to streamline the process. Annotators are also instructed to flag any text pairs they consider faulty or unannotatable. Each annotator is compensated at a rate of approximately \$35 per hour and dedicated between 20 and 35 hours to complete the task.

\begin{figure*}[h]
  \centering
    \includegraphics[width=\textwidth]{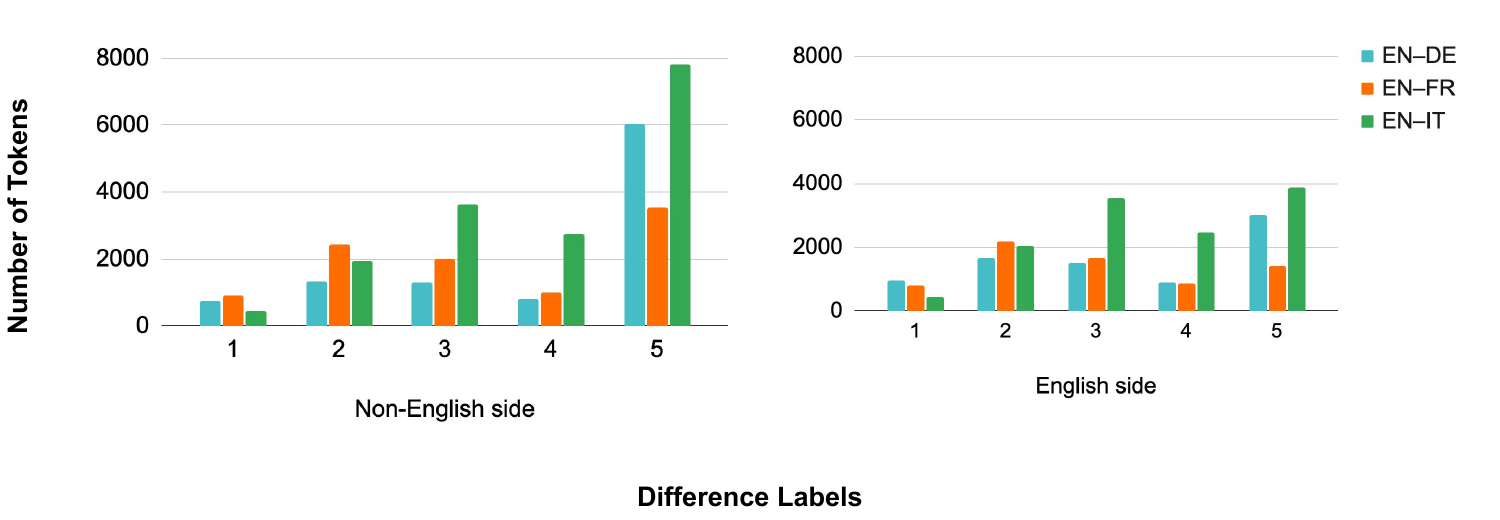}
  \caption{Label distribution of the final SwissGov-RSD dataset in tokens (separated by whitespaces). 0-labeled tokens are not included in this plot.}
  \label{fig:lab-dist}
\end{figure*}

\paragraph{Annotation Validation} We assess annotation quality using automatic agreement metrics computed on the 25 overlapping text pairs annotated by both annotators, $A1$ and $A2$, for each language pair. We define a difference span as a sequence of tokens labeled with the same contiguous label ($>$0). For all such spans $S$, we compute the intersection-over-union (IoU)

\[
\text{IoU} = \frac{|S_{A1} \cap S_{A2}|}{|S_{A1} \cup S_{A2}|},
\]

where the intersection corresponds to all overlapping spans and the union to the total number of annotated difference spans in a document. We report the macro-average over all document IoU scores for each language.
Additionally, we calculate F1 scores between the two annotators.

To evaluate consistency in label assignment for difference spans, we calculate the Spearman rank correlation between the annotators' labels, where non-annotated tokens are excluded from the computation. Correlation is measured for both exact span matches and fuzzy matches, defined as spans with IoU scores larger than 50, 75, and 90 respectively. In addition, to facilitate comparison between annotator agreement and model performance, we report the Spearman correlation over all labels as defined in Subsection~\ref{subsec:eval}. Results are summarized in Table~\ref{tab:main-round}.



While the number of annotated spans varies across language pairs and annotators, IoU and F1 scores remain broadly comparable, suggesting that annotators recognize similar spans despite differing annotation strategies.
Across language pairs, EN--DE shows the lowest label agreement (correlation 76.47 at $\geq$50 overlap) despite comparable overlap statistics with other language pairs, indicating diverging label choices over similar spans. EN--FR presents the inverse pattern: high label correlations (up to 98.05) alongside lower overlap statistics, reflecting stronger consensus on labels than on exact span boundaries. For EN--IT we observe the highest overall overlap agreement and high label correlation.
The inter-annotator Spearman correlation ranges from 43.61 (EN--FR) to 65.09 (EN--IT)\footnote{After a preliminary quality assessment, we identified one below-average annotation run for EN--DE and repeated it with a new annotator. Replacing the annotations from the first run led to an increase in Spearman correlation from 9.4 to 56.14.}. While these values may appear low, they are consistent with agreement reported in previous literature on cross-lingual, pairwise span-level annotation tasks such as error span annotation in translation evaluation \citep{kreutzer-correct, Kocmi2024-wv} and overlap statistics in span-level error detection~\citep{Lavie2025-ri} or quality estimation for post-editing~\citep{Sarti2025-po}, since these tasks are inherently difficult for human annotators as well.  An analysis of the effect of annotation discrepancy  on model evaluation is shown in Appendix~\ref{app:ann-disc}.





We merge the sets of annotated documents. Where documents have been annotated by both annotators, we manually inspect them. We include the annotations adhering more to the guidelines and the label distribution of the other annotators in the dataset. Furthermore, the documents flagged as faulty were also manually inspected and removed if necessary. After filtering 11 faulty documents, the final dataset comprises 224 multi-parallel documents, totaling 886 texts (Table~\ref{tab:data-length}).


 Figure~\ref{fig:lab-dist} shows the label distribution of the dataset. We observe that around 10--17\% of tokens are labeled as semantically different---a considerably more skewed semantic equivalence-to-difference ratio than in the iSTS-RSD dataset with approximately 25--30\%. 

\begin{figure}[h]
    \centering
    \includegraphics[width=0.94\linewidth]{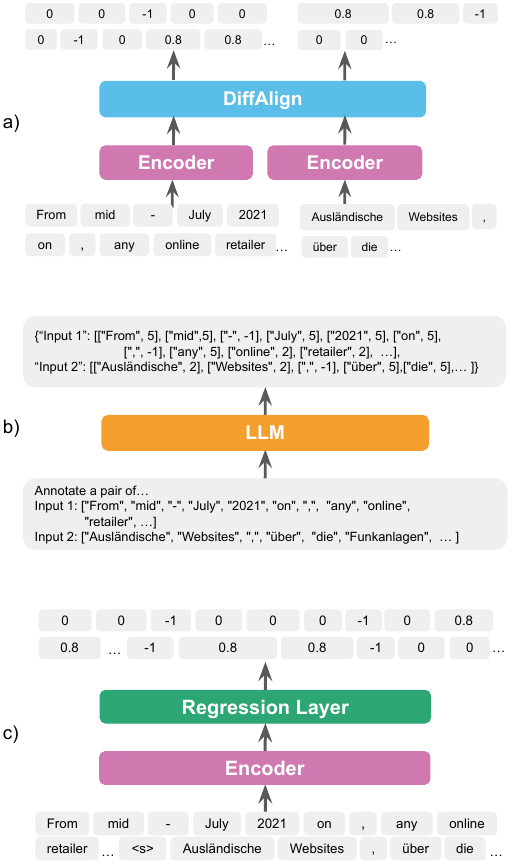}
    \caption{Architectures used in our experiments. An \textbf{unsupervised approach} (a), where each document is encoded separately and a difference alignment algorithm is used to predict difference scores for each token.  We provide \textbf{LLMs} (b) with a natural language instruction and examples of the expected output. The two text segments to compare are provided in tokenized form. The LLM then autoregressively generates a JSON object with a score for each token.
    The \textbf{token regressor} (c) predicts a score for each encoded token in the sequence pair.}
    \label{fig:schem-example}
\end{figure}

\section{Automatic Detection Approaches}

In order to assess how well state-of-the-art methods align with human judgment in recognizing semantic differences, we evaluate three different approaches: an unsupervised RSD algorithm and common specialization techniques, namely few-shot prompting and fine-tuning. 

\subsection{Unsupervised Baseline}
As a baseline, we use DiffAlign, the best-performing RSD approach described by~\citet{vamvas-sennrich-2023-towards} (Figure~\ref{fig:schem-example}a).

Given a pair of texts $A$ and $B$, each is independently encoded into a sequence of contextualized token embeddings $h(A) = [h(a_1), \ldots, h(a_n)]$ and $h(B) = [h(b_1), \ldots, h(b_m)]$. For this we use the final hidden states of a pretrained encoder model. For each token $a_i$ in $A$, a soft alignment score is computed as the maximum cosine similarity between $h(a_i)$ and all token embeddings in $B$:

\[
\textrm{diff}_{align}(a_i) = 1 - \max_{b_j \in B} \cos(h(a_i), h(b_j)).
\]

The resulting score reflects the degree to which a token in one text is semantically aligned with the other. Tokens with low alignment confidence (i.e., high DiffAlign values) are interpreted as likely semantic differences. The method is entirely unsupervised and requires no labeled training data. 

\subsection{Few-shot Prompting}
We instruct an LLM to label every token of a sequence pair according to their semantic similarity on a scale from 0 to 5, equivalent to the labeling scheme employed by iSTS~\cite{agirre-etal-2016-semeval}. The output is to be given in JSON format. We add three in-context examples that illustrate the task and the desired format (Figure~\ref{fig:schem-example}b). The prompt is attached in Appendix~\ref{app:few-shot-prompt}. 

\subsection{Fine-tuning}
\paragraph{LLMs} For fine-tuning LLMs, we use a supervised fine-tuning (SFT) objective that minimizes the cross-entropy between the model’s predictions and the gold response, given the few-shot prompt.

\paragraph{Encoder Models} Here, we concatenate the two input sequences, separated by a special token, and encode them with the model. The final hidden states of the tokens are passed through a linear regression layer to predict a score for each token (Figure~\ref{fig:schem-example}c).
During training, we minimize the binary cross-entropy between the predicted scores and the human-annotated labels. The scores are on a continuous scale from 0 to 1, where 0 indicates perfect alignment and 1 the strongest semantic difference, which is consistent with the scale used by~\citet{vamvas-sennrich-2023-towards}.

\section{Experimental Setup}


\subsection{LLMs}
We evaluate open-weight models of two sizes, Llama-3.1-8B-Instruct and its 405B equivalent~\citep{grattafiori2024llama3herdmodels}, as well as two closed-source commercial models, GPT-4o-mini and GPT-4o~\citep{openai2024gpt4ocard}.
Additionally, we evaluate two models that generate reasoning tokens before predicting the final response: the open-weight DeepSeek-R1~\citep{deepseekai2025deepseekr1incentivizingreasoningcapability} and the closed-source commercial o3-mini~\citep{openai2025competitiveprogramminglargereasoning}, for which we set the reasoning effort to ``low''. 

For Llama, we generate responses using greedy decoding, while for GPT, we use the default settings of the API (i.e., sampling without temperature scaling) and enforce valid JSON output.

For fine-tuning Llama 8B, we train LoRA~\citep{hu2021loralowrankadaptationlarge} as implemented in the PEFT library\footnote{\href{https://github.com/huggingface/peft}{https://github.com/huggingface/peft}} with a rank of $r = 32$ and a scaling factor of $\alpha = 16$, and we use 4-bit quantization during training~\citep{Dettmers2023-hs}. We fine-tune the model with a batch size of 16 and a learning rate of $2 \times 10^{-4}$ for 10 epochs, with early stopping on the validation set. 

To fine-tune GPT-4o-mini, we use the default settings recommended by OpenAI (3 epochs, batch size 1, lr multiplier 1.8).

\subsection{Encoders}
For DiffAlign, we follow~\citet{vamvas-sennrich-2023-towards} and use XLM-R + SimCSE~\citep{conneau-etal-2020-unsupervised, gao-etal-2021-simcse, Wang2022-po}, which has been shown to produce high-quality semantic representations for token-level alignment tasks, and LaBSE~\citep{feng-etal-2022-language}. Since the sequence length of XLM-R and LaBSE is restricted to 512, we also experiment with ModernBERT-large~\citep{warner2024smarterbetterfasterlonger}, its multilingual variants EuroBERT (210M, 610M, 2.1B)~\citep{Boizard2025-rf}, MmBERT~\citep{Marc2025-ea}, as well as bge-m3~\citep{chen-etal-2024-m3}, and Qwen3-Embedding~\citep{Zhang2025-kx}, which allow us to input a whole document without splitting, due to their native sequence length of 8,192 tokens. 

For our encoder fine-tuning experiments, we use ModernBERT-large~\cite{warner2024smarterbetterfasterlonger}. We train the models with a batch size of 32 for 4 epochs, using early stopping based on the validation loss and select the best
model based on validation loss out of four learning rates: $1 \times 10^{-5}$,  $3 \times 10^{-5}$, $5 \times 10^{-5}$, and $8 \times 10^{-5}$. The last learning rate performed best.

Details on the individual LLMs and encoder models are presented in Appendix~\ref{app:description-of-models}.

\subsection{Fine-tuning Data}
For fine-tuning, we use the train splits from iSTS-RSD~\citep{vamvas-sennrich-2023-towards}, which are based on the human-annotated data from SemEval-2016 Task 2~\citep{agirre-etal-2016-semeval-2016}, and 1500 human-validated paraphrases from PAWS~\citep{zhang-etal-2019-paws} to augment negative samples (texts without semantic differences). Label schemes of the latter are converted to the token-level. To train the models on longer inputs beyond single sentences, we apply the same data augmentations used by~\citet{vamvas-sennrich-2023-towards}, concatenating inputs into synthetic ‘documents’ of up to 5 sentences, with up to 5 permutations of the sentence order between document pairs. To cover all language pairs in SwissGov-RSD and enable a direct comparison, we extend iSTS-RSD to EN-IT\footnote{To do so, we use DeepL to machine translate the English iSTS dataset to Italian and re-align the sentence pairs to form an EN-IT dataset: \url{https://huggingface.co/datasets/ZurichNLP/rsd-ists-2016}. In addition, we machine-translate PAWS from English to Italian to also support augmentation with negative examples: \url{https://huggingface.co/datasets/ZurichNLP/paws-x-italian}.}. 

The iSTS-RSD data contains labeled examples only in English, limiting support to English-only fine-tuning. To support cross-lingual fine-tuning without requiring human annotations in the non-English language, we augment the machine-translated side of the training sets with label projection~\citep{akbik-etal-2015-generating, chen-etal-2023-frustratingly}. For each training example, we provide GPT-4o-mini with the English source sentence and its corresponding human-annotated token-level labels, prompting it to assign equivalent labels to the machine-translated version in German, French, or Italian. Prompt templates for each language pair are provided in Appendix~\ref{app:label-projection-prompts}. This approach bypasses the need for explicit translation of span annotations and leverages the LLM’s multilingual and instruction-following capabilities. We refer to this approach as \textit{direct LLM-based label projection} to distinguish it from multi-step or marker-based projection approaches~\citep{chen-etal-2023-frustratingly, le2024constraineddecodingcrosslinguallabel, parekh-etal-2024-contextual}. In Appendix~\ref{app:lp-eval}, we present the results of a manual evaluation of the label projections. For experiments using projected labels, we apply the same data sources and augmentation strategies, with the distinction that one side of each pair is in German, French, or Italian. 

\begin{table*}[t]
\centering
\small
\begin{tabular}{@{}lccc@{\hskip 0.8cm}ccc@{}}
\toprule
\textbf{Approach} & \multicolumn{3}{c}{\textbf{iSTS-RSD}\phantom{000}} & \multicolumn{3}{c}{\textbf{SwissGov-RSD}} \\
\cmidrule(l{0.4cm}r{1cm}){2-4} \cmidrule(lr){5-7}
& \textbf{EN-DE} & \textbf{EN-FR} & \textbf{EN-IT} & \textbf{EN-DE} & \textbf{EN-FR} & \textbf{EN-IT} \\
\midrule
\multicolumn{7}{@{}l@{}}{\textit{DiffAlign (unsupervised)}} \\
XLM-R+SimCSE$^{\dagger}$ 
  & 44.9 & 45.2 & 44.9 
  & 19.2 & 10.0 & 23.4 \\
LaBSE$^{\dagger}$
  & 40.6 & 42.3 & 45.3
  & \textbf{\underline{20.6}} & \phantom{0}8.2 & \textbf{\underline{25.6}} \\
bge-m3 
  & \underline{47.1} & \underline{45.8} & \underline{47.5}
  & 14.8 & \phantom{0}9.9 & 13.6 \\
gte-multilingual-base
  & 31.2 & 30.2 & 30.5
  & 10.7 & \phantom{0}5.9 & 12.8 \\
ModernBERT-large 
  & 17.3 & 16.7 & 17.2 
  & \phantom{0}0.5 & \phantom{0}0.7 & \phantom{0}2.7 \\
EuroBERT 210M  
  & 32.1 & 34.6 & 33.4
  & 11.5 & \phantom{0}8.4 & 16.1 \\
EuroBERT 610M 
  & 29.1 & 30.5 & 31.0
  & 18.6 & 11.4 & 18.7 \\
EuroBERT 2.1B 
  & 40.3 & 28.5 & 28.3
  & 13.7 & \phantom{0}8.9 & 17.4 \\
MmBERT-small 
  & 23.0 & 23.5 & 22.1
  & 10.3 & \phantom{0}5.5 & 11.0 \\
MmBERT-base 
  & 34.0 & 32.6 & 32.6
  & 14.3 & \phantom{0}8.6 & 15.6 \\
Qwen3-Embedding 4B  
  & 41.6 & 42.9 & 43.2
  & 20.1 & 11.2 & 20.3 \\
Qwen3-Embedding 8B 
  & 40.9 & 40.1 & 41.4
  & 19.3 & \textbf{\underline{12.1}} & 18.6 \\
\midrule
\multicolumn{7}{@{}l@{}}{\textit{LLMs with few-shot prompting}} \\
Llama-3.1 8B Instruct & \phantom{0}2.8 & \phantom{0}3.7 & \phantom{0}2.6 & - & - & - \\
Llama-3.1 405B Instruct & 29.2 & 29.3 & 26.9 & -0.8 & -2.2 & \phantom{0}\underline{7.4} \\
GPT-4o-mini & 15.8 & 13.8 & 16.6 & - & - & - \\
GPT-4o & 43.0 & 40.5 & 42.5 & \phantom{0}\underline{4.2} & \phantom{0}\underline{1.2} & \phantom{0}5.5 \\
DeepSeek-R1 & 38.2 & 40.1 & 34.6 & - & - & - \\
o3-mini-low & \underline{44.8} & \underline{46.5} & \underline{48.2} & - & - & - \\
\midrule
\multicolumn{7}{@{}l@{}}{\textit{Fine-tuned LLMs}} \\
Llama-3.1 8B Instruct & 66.7 & 67.3 & 66.9 & - & - & - \\
GPT-4o-mini & \textbf{\underline{81.6}} & \textbf{\underline{79.9}} & \textbf{\underline{78.2}} & \phantom{0}\underline{3.9} & \phantom{0}\underline{1.1} & \phantom{0}\underline{5.3} \\
\midrule
\multicolumn{7}{@{}l@{}}{\textit{Fine-tuned encoder models}} \\
ModernBERT-large (EN) & 55.3 & 55.4 & 53.8 & \phantom{0}5.7 & \phantom{0}1.3 & \phantom{0}4.0 \\
ModernBERT-large (EN-DE*) & 58.8 & 58.4 & 58.5 & \phantom{0}6.4 & \phantom{0}1.6 & \phantom{0}5.6 \\
ModernBERT-large (EN-FR*) & 58.4 & 60.7 & 60.1 &  \phantom{0}5.4 & \phantom{0}\underline{5.2} & \phantom{0}8.7  \\
ModernBERT-large (EN-IT*) & \underline{59.2} & \underline{61.2} & \underline{63.0} & \phantom{0}\underline{7.5} & \phantom{0}4.8 & \underline{12.4} \\
ModernBERT-large (multi*) & 54.7 & 57.5 & 58.1 & \phantom{0}5.7 & \phantom{0}2.5 & \phantom{0}8.8 \\
\bottomrule
\end{tabular}
\caption{Token-level Spearman correlations with gold labels for three language pairs, comparing results on the iSTS-RSD and the SwissGov-RSD dataset. (*) denotes encoders fine-tuned on data with projected labels, \textbf{bold} the best performance overall, and \underline{underline} best performance within model category. ($^{\dagger}$) denotes that the input sequences for SwissGov had to be split due to length limitations of the model. Due to computational restraints, for LLM-based experiments, we only consider the best-performing systems during iSTS-RSD evaluation for SwissGov as well.}
\label{tab:main-results}
\end{table*}
This approach allows us to invest comparable amounts of annotated data during fine-tuning across all models, while adjusting the number of augmentations based on computational resources: For fine-tuning the LLMs, which have more parameters and require more computational resources, we use 560 augmentations for training. For the encoder models, we generate 10,000 augmentations. For the multilingually fine-tuned encoder, we concatenate the EN--DE, EN--FR, and EN--IT training data with projected labels. Detailed statistics of the iSTS-RSD training and test sets are shown in Appendix~\ref{app:description-of-datasets} and an example of an augmented sample is presented in Appendix~\ref{app:augmented_input}.

\subsection{Evaluation}
\label{subsec:eval}
We evaluate the above-described systems on the test split of iSTS-RSD as well as on the whole SwissGov-RSD dataset. 

The test split of the iSTS-RSD benchmark includes 5 difficulty settings ranging from individual English sentence pairs (‘iSTS’) to a challenging cross-lingual setting where documents are sequences of 5 permuted sentences including 7 different language pairs: EN--DE, EN--FR, EN--IT, EN--ES, EN--ZH, EN--JA, and EN--KO. We restrict the evaluation to 100 samples per difficulty setting and language to reduce computational cost. 

Some LLM-generated outputs were not in the expected format, either due to additional text outside the expected JSON format or a mismatch in the number of token-label pairs. In the former case, we apply heuristics to extract the JSON object; in the latter, we pad or truncate the output to align with the gold sequence length to enable evaluation. Furthermore, the LLM-generated output sometimes contains labels outside the expected range. These were not altered and used as is for correlation computation. 

To evaluate model performance, we use token-level Spearman correlations between all gold labels and predictions, where semantically equivalent tokens are labeled 0. Since Spearman correlations can be unstable if there are many ties, we also compute Kendall $\tau$-b scores for the results in Table~\ref{tab:main-results} and present them in Appendix~\ref{app:kendall-results}, showing no substantial differences from Spearman correlation-based evaluation.

\begin{figure*}[t]
  \centering
  \includegraphics[width=\textwidth]{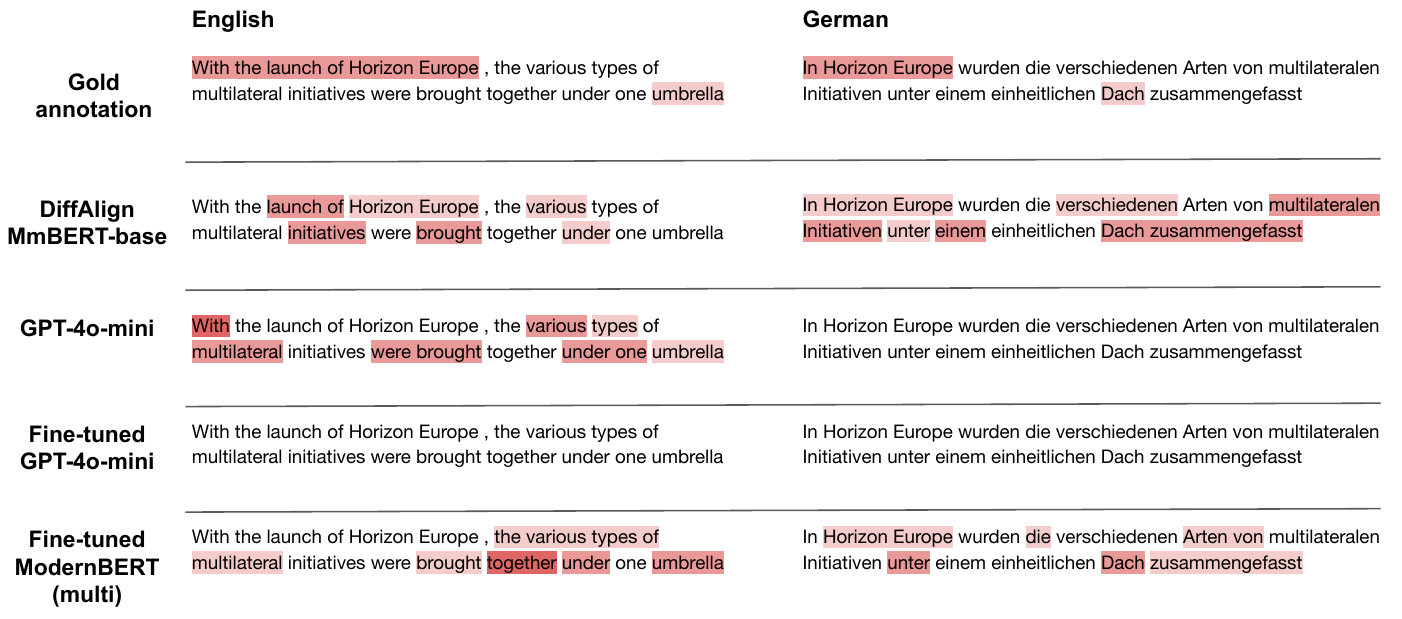}
  \caption{Excerpt of an EN--DE document pair with gold labels and predictions by one model from each of the system categories listed in Table~\ref{tab:main-results}.}
\label{fig:pred-example}
\end{figure*}

\section{Results and Discussion}

\paragraph{Synthetically augmented datasets do not sufficiently reflect properties of a realistic scenario.} Table~\ref{tab:main-results} presents a comparison of model performance on the EN--DE, EN--FR, EN--IT subsets of iSTS-RSD\footnote{A comprehensive overview of the results over the different augmentation categories and languages of iSTS-RSD are presented in Appendix~\ref{app:ists}.} and the full SwissGov-RSD benchmark. Most notably, across all model types and language pairs, performance drops substantially when moving from iSTS-RSD to SwissGov-RSD. This confirms that models tuned or evaluated only on synthetically augmented data do not generalize well to naturalistic, real-world scenarios. The fine-tuned GPT-4o-mini model exemplifies this gap the most: Despite achieving the highest scores on iSTS-RSD, its performance drops to among the lowest on SwissGov-RSD, suggesting potential overfitting to the data augmentation patterns. 
Interestingly, while overall scores on SwissGov-RSD are lower, the unsupervised approaches show relatively strong robustness, outperforming all other few-shot and fine-tuned models. This may indicate that more specialized models suffer from out-of-domain effects, as SwissGov-RSD differs substantially from iSTS-RSD not only in terms of naturalness, but also in domain, length, and label distribution.

\paragraph{Encoders are competitive.} Encoder-based models remain highly competitive, particularly in unsupervised settings (Table~\ref{tab:main-results}).
Among long-context models, Qwen3-Embedding models achieve the strongest results, followed by EuroBERT 610M.
While LaBSE and XLM-R + SimCSE achieve the highest scores in the unsupervised setting, their limited input length requires document segmentation prior to processing. This effectively reduces the task to shorter units and avoids challenges associated with long-context modeling. The higher performance under segmentation suggests that long-context processing, even when supported, may weaken token-level cross-lingual alignment signals necessary for RSD.
DiffAlign underperforms when using ModernBERT, likely due to its English-centricity. However, when fine-tuned, it surpasses LLMs in a naturalistic setting with few-shot prompting, all while being more efficient in terms of time and computational resources (see Appendix~\ref{app:inference-time} for an inference time comparison).

\begin{table}[t]
\centering
\small
\begin{tabular}{@{}lrr@{}}
\toprule
\textbf{Approach} & \textbf{iSTS-RSD} & \textbf{SwissGov-RSD} \\
\midrule
\multicolumn{3}{@{}l@{}}{\textit{LLMs with few-shot prompting}} \\
Llama-3.1 8B Instruct & 0.60\% & -- \\
Llama-3.1 405B Instruct & 0.07\% & 7.00\% \\
GPT-4o & 0.03\% & 2.10\% \\
\midrule
\multicolumn{3}{@{}l@{}}{\textit{Fine-tuned LLMs}} \\
Llama-3.1 8B Instruct & 0.10\% & -- \\
GPT-4o-mini & 0.07\% & 7.40\% \\
\bottomrule
\end{tabular}
\caption{Percentage of samples for which LLMs failed to produce the correct number of labels for the three investigated language pairs (see Table~\ref{tab:llm-fail-full} for full iSTS-RSD). For SwissGov-RSD, each text of a pair is counted separately, hence, the model can fail twice for one sample.}
\label{tab:llm-fail}
\end{table}

\begin{figure}[t]
  \centering
  \includegraphics[width=\linewidth]{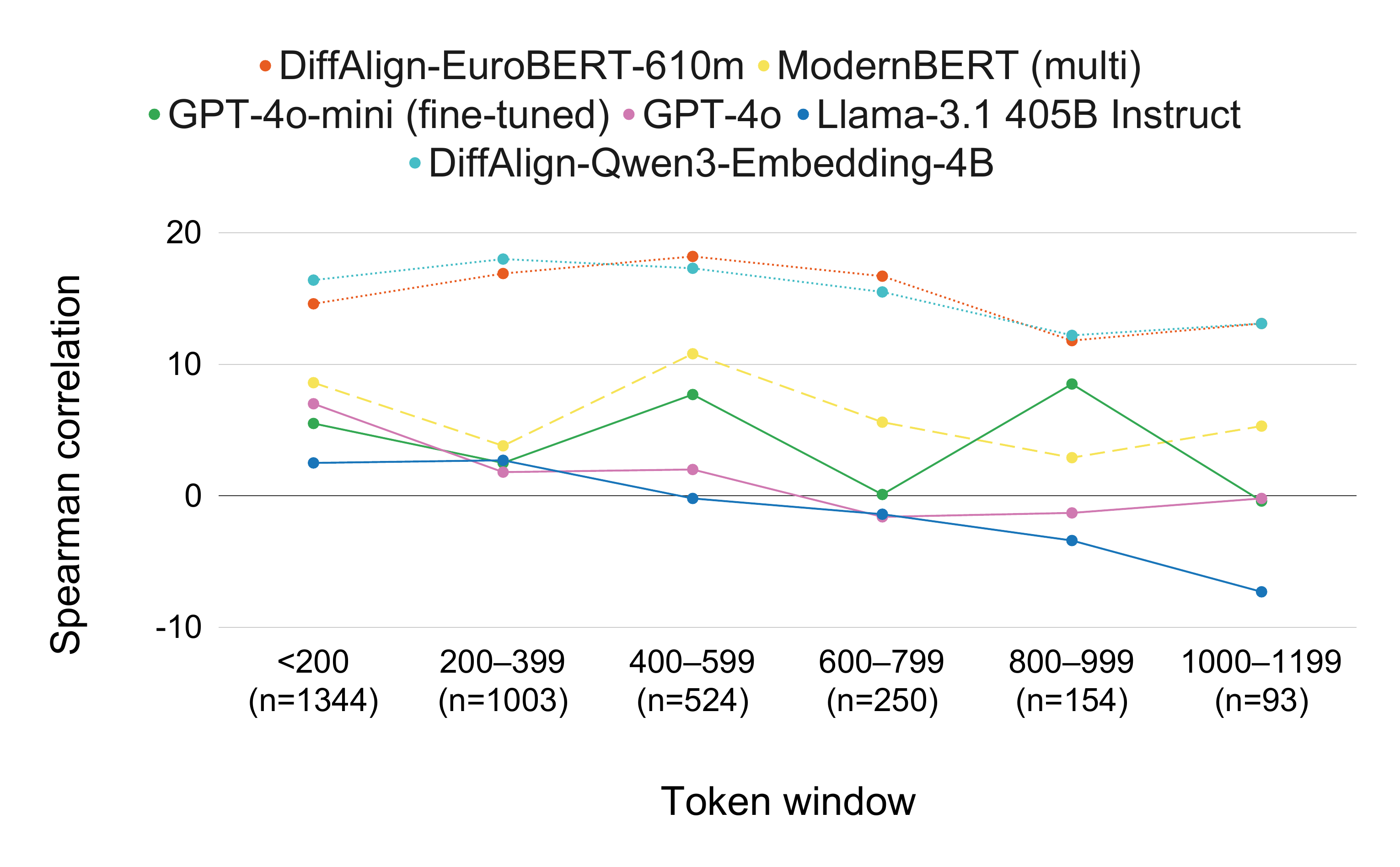}
  \caption{Average Spearman correlation coefficient at different positions in the documents, averaged across language pairs. $n$ indicates the number of documents. Dotted lines represent DiffAlign approaches, dashed the fine-tuned regressor, and solid LLMs.}
  \label{fig:position-corr}
\end{figure}


\paragraph{LLMs are highly limited on this task.} Table~\ref{tab:main-results} shows that for iSTS-RSD, fine-tuned GPT-4o-mini performs best, while few-shot prompted LLMs fail to beat the unsupervised baseline, with reasoning bringing only little improvement. On SwissGov-RSD, however, LLMs show the largest performance drops, especially fine-tuned GPT-4o-mini. Proprietary models generally outperform open-weight alternatives, and despite strong performance in synthetically augmented settings, LLMs are prone to output formatting errors, sequence length mismatches in particular, which worsen on SwissGov-RSD (Table~\ref{tab:llm-fail}), a problem encoder-based models avoid by design. Qualitatively, LLMs also tend to underlabel semantic differences, most noticeably toward document ends or in the second input document, as illustrated in Figure~\ref{fig:pred-example}.

\paragraph{Long documents expose weaknesses.} To better understand how models behave on extended inputs, we analyze the correlation between model predictions and gold labels by token position in Figure~\ref{fig:position-corr}. While all systems show a slight downward trend as documents grow longer,  encoder-based systems, specifically the unsupervised DiffAlign approach, appear more robust than LLMs. The plot suggests that LlaMA-3.1 405B is most sensitive to document length, often failing to produce complete output for inputs exceeding 1200 tokens. This suggests that unsupervised approaches with encoder models or in-domain fine-tuning may be promising strategies for cross-lingual RSD for long documents. For LLMs, however, future work may need to investigate strategies that preserve structured output quality for sequence labeling at extended input lengths.

\section{Related Work}

RSD is closely related to STS, paraphrase detection, and natural language inference (NLI). Apart from the above-mentioned \mbox{(i-)STS(-RSD)} and \mbox{PAWS(-X)} datasets~\citep{agirre-etal-2016-semeval, agirre-etal-2016-semeval-2016, vamvas-sennrich-2023-towards, zhang-etal-2019-paws, yang-etal-2019-paws}, standard NLI datasets like SNLI, MNLI, and XNLI~\citep{bowman-etal-2015-large, williams-etal-2018-broad, conneau-etal-2018-xnli} also relate, but provide sentence-level annotations, whereas our task operates at token-level granularity.

RSD also intersects with evaluation of generated text, particularly in machine translation and hallucination detection. Datasets using the MQM framework~\citep{lommel-etal-2024-multi}, such as MQM~\citep{freitag-etal-2021-experts}, QT~\citep{specia-etal-2017-translation}, and ACES~\citep{amrhein-etal-2022-aces, moghe-etal-2025-machine}, annotate translation errors which can also include semantic differences. Hallucination detection benchmarks, such as HaDes~\citep{liu-etal-2022-token}, 
and multilingual Mushroom~\citep{mickus-etal-2024-semeval, vázquez2025semeval2025task3mushroom} include semantic errors but are not parallel.

Our dataset also relates to comparable corpora, which contain semantically aligned but not fully parallel documents. Related corpora include \mbox{SwissAdmin}~\citep{scherrer-etal-2014-swissadmin}, the Bulletin Corpus~\citep{inproceedings}, and 20min-XD~\citep{wastl-etal-2025-20min}, which similarly leverage Swiss multilingual content but do not include token-level semantic annotations.

\section{Conclusion}

We release the first human-annotated, cross-lingual, document-level dataset consisting of naturally occurring semantic differences in multilingual government texts: SwissGov-RSD. Through a comprehensive evaluation of state-of-the-art systems, including LLMs, encoder models, and widely used specialization techniques on SwissGov-RSD and previous synthetically augmented benchmarks, we find a considerable performance gap between the synthetic and naturalistic evaluation settings. Our findings indicate that current systems struggle to generalize to real-world data, as data augmentation fails to capture the complexity of naturalistic settings.
This emphasizes that document-level RSD across languages remains a complex challenge calling for new approaches that align better with human judgment and are applicable in real-world scenarios.

\section*{Limitations}

\paragraph{Language and script diversity} SwissGov-RSD is limited to high-resource, closely related languages (German, French, Italian) and Latin scripts. Furthermore, the language pairs chosen for the dataset are English-centric. While this setting should be comparatively favorable, our results show that current models already struggle under these conditions. At the same time, the generalizability of our findings to typologically more distant languages and those using non-Latin scripts remains untested in a realistic setting. Due to the dataset's multi-parallel nature, future work could extend the annotation to other language combinations, although we note the high annotation cost.

\paragraph{Annotation consistency} Annotation quality may vary across annotators. Ideally, multiple annotators per document pair for all samples would be used to ensure high reliability. Due to resource constraints, we relied on a two-annotator setup with overlap-based quality checks.

\paragraph{Prompting simplicity} Our prompting approach is intentionally simple and uniform across models to support comparability. While more elaborate prompts might improve performance marginally, the primary issues we observe suggest deeper limitations in LLM robustness that prompt tuning alone is unlikely to resolve.

\section*{Acknowledgements}
This work was funded by the Swiss National Science Foundation (project InvestigaDiff; no.~10000503). We thank the reviewers for their constructive feedback and valuable suggestions, and are grateful to Cui Ding and Zach Hopton for proofreading.

\bibliography{anthology,custom}

@inproceedings{zhang2024mgte, title={mGTE: Generalized Long-Context Text Representation and Reranking Models for Multilingual Text Retrieval}, author={Zhang, Xin and Zhang, Yanzhao and Long, Dingkun and Xie, Wen and Dai, Ziqi and Tang, Jialong and Lin, Huan and Yang, Baosong and Xie, Pengjun and Huang, Fei and others}, booktitle={Proceedings of the 2024 Conference on Empirical Methods in Natural Language Processing: Industry Track}, pages={1393--1412}, year={2024} }

@inproceedings{chen-etal-2024-m3, title = "{M}3-Embedding: Multi-Linguality, Multi-Functionality, Multi-Granularity Text Embeddings Through Self-Knowledge Distillation", author = "Chen, Jianlyu and Xiao, Shitao and Zhang, Peitian and Luo, Kun and Lian, Defu and Liu, Zheng", editor = "Ku, Lun-Wei and Martins, Andre and Srikumar, Vivek", booktitle = "Findings of the Association for Computational Linguistics: ACL 2024", month = aug, year = "2024", address = "Bangkok, Thailand", publisher = "Association for Computational Linguistics", url = "https://aclanthology.org/2024.findings-acl.137/", doi = "10.18653/v1/2024.findings-acl.137", pages = "2318--2335" }

@ARTICLE{Zhang2025-kx, title = "{Qwen3} embedding: Advancing text embedding and reranking through foundation models", author = "Zhang, Yanzhao and Li, Mingxin and Long, Dingkun and Zhang, Xin and Lin, Huan and Yang, Baosong and Xie, Pengjun and Yang, An and Liu, Dayiheng and Lin, Junyang and Huang, Fei and Zhou, Jingren", journal = "arXiv [cs.CL]", month = jun, year = 2025, archivePrefix = "arXiv", primaryClass = "cs.CL" }

@ARTICLE{Marc2025-ea, title = "{MmBERT}: A modern multilingual encoder with annealed language learning", author = "Marc, Marone and Orion, Weller and William, Fleshman and Eugene, Yang and Dawn, Lawrie and Van Durme, Benjamin", journal = "arXiv [cs.CL]", month = sep, year = 2025, archivePrefix = "arXiv", primaryClass = "cs.CL" }

@INPROCEEDINGS{Lavie2025-ri,
  title     = "Findings of the {WMT25} shared task on automated translation
               evaluation systems: Linguistic diversity is challenging and
               references still help",
  author    = "Lavie, Alon and Hanneman, Greg and Agrawal, Sweta and Kanojia,
               Diptesh and Lo, Chi-Kiu and Zouhar, Vilém and Blain, Frederic and
               Zerva, Chrysoula and Avramidis, Eleftherios and Deoghare, Sourabh
               and Sindhujan, Archchana and Wang, Jiayi and Adelani, David
               Ifeoluwa and Thompson, Brian and Kocmi, Tom and Freitag, Markus
               and Deutsch, Daniel",
  booktitle = "Proceedings of the Tenth Conference on Machine Translation",
  publisher = "Association for Computational Linguistics",
  address   = "Stroudsburg, PA, USA",
  pages     = "436--483",
  year      =  2025,
  language  = "en"
}

@inproceedings{ Boizard2025-rf, title={Euro{BERT}: Scaling Multilingual Encoders for European Languages}, author={Nicolas Boizard and Hippolyte Gisserot-Boukhlef and Duarte Miguel Alves and Andre Martins and Ayoub Hammal and Caio Corro and Celine Hudelot and Emmanuel Malherbe and Etienne Malaboeuf and Fanny Jourdan and Gabriel Hautreux and Jo{\~a}o Alves and Kevin El Haddad and Manuel Faysse and Maxime Peyrard and Nuno M Guerreiro and Patrick Fernandes and Ricardo Rei and Pierre Colombo}, booktitle={Second Conference on Language Modeling}, year={2025}, url={https://openreview.net/forum?id=jdOC24msVq} }

@INPROCEEDINGS{Kocmi2024-wv, title = "Error span annotation: A balanced approach for human evaluation of machine translation", author = "Kocmi, Tom and Zouhar, Vilém and Avramidis, Eleftherios and Grundkiewicz, Roman and Karpinska, Marzena and Popović, Maja and Sachan, Mrinmaya and Shmatova, Mariya", booktitle = "Proceedings of the Ninth Conference on Machine Translation", publisher = "Association for Computational Linguistics", address = "Stroudsburg, PA, USA", pages = "1440--1453", month = nov, year = 2024 }

@INPROCEEDINGS{Wang2022-po, title = "English contrastive learning can learn universal cross-lingual sentence embeddings", author = "Wang, Yaushian and Wu, Ashley and Neubig, Graham", booktitle = "Proceedings of the 2022 Conference on Empirical Methods in Natural Language Processing", publisher = "Association for Computational Linguistics", address = "Stroudsburg, PA, USA", pages = "9122--9133", month = dec, year = 2022 }

@inproceedings{kreutzer-correct,
    title = "Correct Me If You Can: Learning from Error Corrections and Markings",
    author = "Kreutzer, Julia  and
      Berger, Nathaniel  and
      Riezler, Stefan",
    editor = "Martins, Andr{\'e}  and
      Moniz, Helena  and
      Fumega, Sara  and
      Martins, Bruno  and
      Batista, Fernando  and
      Coheur, Luisa  and
      Parra, Carla  and
      Trancoso, Isabel  and
      Turchi, Marco  and
      Bisazza, Arianna  and
      Moorkens, Joss  and
      Guerberof, Ana  and
      Nurminen, Mary  and
      Marg, Lena  and
      Forcada, Mikel L.",
    booktitle = "Proceedings of the 22nd Annual Conference of the European Association for Machine Translation",
    month = nov,
    year = "2020",
    address = "Lisboa, Portugal",
    publisher = "European Association for Machine Translation",
    url = "https://aclanthology.org/2020.eamt-1.15/",
    pages = "135--144"
}

@INPROCEEDINGS{Sarti2025-po, title = "Unsupervised Word-level Quality Estimation for Machine Translation Through the Lens of Annotators (Dis)agreement", author = "Sarti, Gabriele and Zouhar, Vilém and Nissim, Malvina and Bisazza, Arianna", booktitle = "Proceedings of the 2025 Conference on Empirical Methods in Natural Language Processing", pages = "18320--18337", month = nov, year = 2025 }

@INPROCEEDINGS{inproceedings, title = "Building a Parallel Corpus on the World's Oldest Banking Magazine", author = "Volk, Martin and Amrhein, Chantal and Aepli, Noëmi and Müller, Mathias and Ströbel, Phillip", booktitle = "KONVENS", publisher = "s.n.", month = sep, year = 2016, language = "en" }

@inproceedings{wastl-etal-2025-20min,
    title = "20min-{XD}: A Comparable Corpus of {S}wiss News Articles",
    author = "Wastl, Michelle  and
      Vamvas, Jannis  and
      Calleri, Selena  and
      Sennrich, Rico",
    editor = {Gerber, Jonathan  and
      Cieliebak, Mark  and
      Tuggener, Don  and
      H{\"u}rlimann, Manuela},
    booktitle = "Proceedings of the 10th edition of the Swiss Text Analytics Conference",
    month = may,
    year = "2025",
    address = "Winterthur, Switzerland",
    publisher = "Association for Computational Linguistics",
    url = "https://aclanthology.org/2025.swisstext-1.1/",
    pages = "1--10"
}

@inproceedings{vázquez2025semeval2025task3mushroom, title = "{S}em{E}val-2025 Task 3: Mu-{SHROOM}, the Multilingual Shared-task on Hallucinations and Related Observable Overgeneration Mistakes", author = {Vazquez, Raul and Mickus, Timothee and Zosa, Elaine and Vahtola, Teemu and Tiedemann, J{\"o}rg and Sinha, Aman and Segonne, Vincent and Sanchez - Vega, Fernando and Raganato, Alessandro and Libovick{\'y}, Jind{\v{r}}ich and Karlgren, Jussi and Ji, Shaoxiong and Helcl, Jind{\v{r}}ich and Guillou, Liane and De Gibert, Ona and Bengoetxea, Jaione and Attieh, Joseph and Apidianaki, Marianna}, editor = "Rosenthal, Sara and Ros{\'a}, Aiala and Ghosh, Debanjan and Zampieri, Marcos", booktitle = "Proceedings of the 19th International Workshop on Semantic Evaluation (SemEval-2025)", month = jul, year = "2025", address = "Vienna, Austria", publisher = "Association for Computational Linguistics", url = "https://aclanthology.org/2025.semeval-1.322/", pages = "2472--2497", ISBN = "979-8-89176-273-2" }

@inproceedings{amrhein-etal-2022-aces, title = "{ACES}: Translation Accuracy Challenge Sets for Evaluating Machine Translation Metrics", author = "Amrhein, Chantal and Moghe, Nikita and Guillou, Liane", editor = {Koehn, Philipp and Barrault, Lo{\"i}c and Bojar, Ond{\v{r}}ej and Bougares, Fethi and Chatterjee, Rajen and Costa-juss{\a}, Marta R. and Federmann, Christian and Fishel, Mark and Fraser, Alexander and Freitag, Markus and Graham, Yvette and Grundkiewicz, Roman and Guzman, Paco and Haddow, Barry and Huck, Matthias and Jimeno Yepes, Antonio and Kocmi, Tom and Martins, Andr{\'e} and Morishita, Makoto and Monz, Christof and Nagata, Masaaki and Nakazawa, Toshiaki and Negri, Matteo and N{\'e}v{\'e}ol, Aur{\'e}lie and Neves, Mariana and Popel, Martin and Turchi, Marco and Zampieri, Marcos}, booktitle = "Proceedings of the Seventh Conference on Machine Translation (WMT)", month = dec, year = "2022", address = "Abu Dhabi, United Arab Emirates (Hybrid)", publisher = "Association for Computational Linguistics", url = "https://aclanthology.org/2022.wmt-1.44/", pages = "479--513" }

@INPROCEEDINGS{Dettmers2023-hs,
  title     = "{QLoRA}: Efficient Finetuning of Quantized {LLMs}",
  author    = "Dettmers, Tim and Pagnoni, Artidoro and Holtzman, Ari and
               Zettlemoyer, Luke",
  booktitle = "Thirty-seventh Conference on Neural Information Processing
               Systems",
  month     =  nov,
  year      =  2023
}

@inproceedings{mickus-etal-2024-semeval, title = "{S}em{E}val-2024 Task 6: {SHROOM}, a Shared-task on Hallucinations and Related Observable Overgeneration Mistakes", author = {Mickus, Timothee and Zosa, Elaine and Vazquez, Raul and Vahtola, Teemu and Tiedemann, J{\"o}rg and Segonne, Vincent and Raganato, Alessandro and Apidianaki, Marianna}, editor = {Ojha, Atul Kr. and Do{\u{g}}ru{\"o}z, A. Seza and Tayyar Madabushi, Harish and Da San Martino, Giovanni and Rosenthal, Sara and Ros{\'a}, Aiala}, booktitle = "Proceedings of the 18th International Workshop on Semantic Evaluation (SemEval-2024)", month = jun, year = "2024", address = "Mexico City, Mexico", publisher = "Association for Computational Linguistics", url = "https://aclanthology.org/2024.semeval-1.273/", doi = "10.18653/v1/2024.semeval-1.273", pages = "1979--1993" }

@inproceedings{lommel-etal-2024-multi, title = "The Multi-Range Theory of Translation Quality Measurement: {MQM} scoring models and Statistical Quality Control", author = "Lommel, Arle and Gladkoff, Serge and Melby, Alan and Wright, Sue Ellen and Strandvik, Ingemar and Gasova, Katerina and Vaasa, Angelika and Benzo, Andy and Marazzato Sparano, Romina and Foresi, Monica and Innis, Johani and Han, Lifeng and Nenadic, Goran", editor = "Martindale, Marianna and Campbell, Janice and Savenkov, Konstantin and Goel, Shivali", booktitle = "Proceedings of the 16th Conference of the Association for Machine Translation in the Americas (Volume 2: Presentations)", month = sep, year = "2024", address = "Chicago, USA", publisher = "Association for Machine Translation in the Americas", url = "https://aclanthology.org/2024.amta-presentations.6/", pages = "75--94" }

@inproceedings{rei-etal-2023-inside, title = "The Inside Story: Towards Better Understanding of Machine Translation Neural Evaluation Metrics", author = "Rei, Ricardo and Guerreiro, Nuno M. and Treviso, Marcos and Coheur, Luisa and Lavie, Alon and Martins, Andr{\'e}", editor = "Rogers, Anna and Boyd-Graber, Jordan and Okazaki, Naoaki", booktitle = "Proceedings of the 61st Annual Meeting of the Association for Computational Linguistics (Volume 2: Short Papers)", month = jul, year = "2023", address = "Toronto, Canada", publisher = "Association for Computational Linguistics", url = "https://aclanthology.org/2023.acl-short.94/", doi = "10.18653/v1/2023.acl-short.94", pages = "1089--1105" }

@misc{openai2024gpt4ocard, title={GPT-4o System Card}, author={OpenAI and : and Aaron Hurst and Adam Lerer and Adam P. Goucher and Adam Perelman and Aditya Ramesh and Aidan Clark and AJ Ostrow and Akila Welihinda and Alan Hayes and Alec Radford and Aleksander Mądry and Alex Baker-Whitcomb and Alex Beutel and Alex Borzunov and Alex Carney and Alex Chow and Alex Kirillov and Alex Nichol and Alex Paino and Alex Renzin and Alex Tachard Passos and Alexander Kirillov and Alexi Christakis and Alexis Conneau and Ali Kamali and Allan Jabri and Allison Moyer and Allison Tam and Amadou Crookes and Amin Tootoochian and Amin Tootoonchian and Ananya Kumar and Andrea Vallone and Andrej Karpathy and Andrew Braunstein and Andrew Cann and Andrew Codispoti and Andrew Galu and Andrew Kondrich and Andrew Tulloch and Andrey Mishchenko and Angela Baek and Angela Jiang and Antoine Pelisse and Antonia Woodford and Anuj Gosalia and Arka Dhar and Ashley Pantuliano and Avi Nayak and Avital Oliver and Barret Zoph and Behrooz Ghorbani and Ben Leimberger and Ben Rossen and Ben Sokolowsky and Ben Wang and Benjamin Zweig and Beth Hoover and Blake Samic and Bob McGrew and Bobby Spero and Bogo Giertler and Bowen Cheng and Brad Lightcap and Brandon Walkin and Brendan Quinn and Brian Guarraci and Brian Hsu and Bright Kellogg and Brydon Eastman and Camillo Lugaresi and Carroll Wainwright and Cary Bassin and Cary Hudson and Casey Chu and Chad Nelson and Chak Li and Chan Jun Shern and Channing Conger and Charlotte Barette and Chelsea Voss and Chen Ding and Cheng Lu and Chong Zhang and Chris Beaumont and Chris Hallacy and Chris Koch and Christian Gibson and Christina Kim and Christine Choi and Christine McLeavey and Christopher Hesse and Claudia Fischer and Clemens Winter and Coley Czarnecki and Colin Jarvis and Colin Wei and Constantin Koumouzelis and Dane Sherburn and Daniel Kappler and Daniel Levin and Daniel Levy and David Carr and David Farhi and David Mely and David Robinson and David Sasaki and Denny Jin and Dev Valladares and Dimitris Tsipras and Doug Li and Duc Phong Nguyen and Duncan Findlay and Edede Oiwoh and Edmund Wong and Ehsan Asdar and Elizabeth Proehl and Elizabeth Yang and Eric Antonow and Eric Kramer and Eric Peterson and Eric Sigler and Eric Wallace and Eugene Brevdo and Evan Mays and Farzad Khorasani and Felipe Petroski Such and Filippo Raso and Francis Zhang and Fred von Lohmann and Freddie Sulit and Gabriel Goh and Gene Oden and Geoff Salmon and Giulio Starace and Greg Brockman and Hadi Salman and Haiming Bao and Haitang Hu and Hannah Wong and Haoyu Wang and Heather Schmidt and Heather Whitney and Heewoo Jun and Hendrik Kirchner and Henrique Ponde de Oliveira Pinto and Hongyu Ren and Huiwen Chang and Hyung Won Chung and Ian Kivlichan and Ian O'Connell and Ian O'Connell and Ian Osband and Ian Silber and Ian Sohl and Ibrahim Okuyucu and Ikai Lan and Ilya Kostrikov and Ilya Sutskever and Ingmar Kanitscheider and Ishaan Gulrajani and Jacob Coxon and Jacob Menick and Jakub Pachocki and James Aung and James Betker and James Crooks and James Lennon and Jamie Kiros and Jan Leike and Jane Park and Jason Kwon and Jason Phang and Jason Teplitz and Jason Wei and Jason Wolfe and Jay Chen and Jeff Harris and Jenia Varavva and Jessica Gan Lee and Jessica Shieh and Ji Lin and Jiahui Yu and Jiayi Weng and Jie Tang and Jieqi Yu and Joanne Jang and Joaquin Quinonero Candela and Joe Beutler and Joe Landers and Joel Parish and Johannes Heidecke and John Schulman and Jonathan Lachman and Jonathan McKay and Jonathan Uesato and Jonathan Ward and Jong Wook Kim and Joost Huizinga and Jordan Sitkin and Jos Kraaijeveld and Josh Gross and Josh Kaplan and Josh Snyder and Joshua Achiam and Joy Jiao and Joyce Lee and Juntang Zhuang and Justyn Harriman and Kai Fricke and Kai Hayashi and Karan Singhal and Katy Shi and Kavin Karthik and Kayla Wood and Kendra Rimbach and Kenny Hsu and Kenny Nguyen and Keren Gu-Lemberg and Kevin Button and Kevin Liu and Kiel Howe and Krithika Muthukumar and Kyle Luther and Lama Ahmad and Larry Kai and Lauren Itow and Lauren Workman and Leher Pathak and Leo Chen and Li Jing and Lia Guy and Liam Fedus and Liang Zhou and Lien Mamitsuka and Lilian Weng and Lindsay McCallum and Lindsey Held and Long Ouyang and Louis Feuvrier and Lu Zhang and Lukas Kondraciuk and Lukasz Kaiser and Luke Hewitt and Luke Metz and Lyric Doshi and Mada Aflak and Maddie Simens and Madelaine Boyd and Madeleine Thompson and Marat Dukhan and Mark Chen and Mark Gray and Mark Hudnall and Marvin Zhang and Marwan Aljubeh and Mateusz Litwin and Matthew Zeng and Max Johnson and Maya Shetty and Mayank Gupta and Meghan Shah and Mehmet Yatbaz and Meng Jia Yang and Mengchao Zhong and Mia Glaese and Mianna Chen and Michael Janner and Michael Lampe and Michael Petrov and Michael Wu and Michele Wang and Michelle Fradin and Michelle Pokrass and Miguel Castro and Miguel Oom Temudo de Castro and Mikhail Pavlov and Miles Brundage and Miles Wang and Minal Khan and Mira Murati and Mo Bavarian and Molly Lin and Murat Yesildal and Nacho Soto and Natalia Gimelshein and Natalie Cone and Natalie Staudacher and Natalie Summers and Natan LaFontaine and Neil Chowdhury and Nick Ryder and Nick Stathas and Nick Turley and Nik Tezak and Niko Felix and Nithanth Kudige and Nitish Keskar and Noah Deutsch and Noel Bundick and Nora Puckett and Ofir Nachum and Ola Okelola and Oleg Boiko and Oleg Murk and Oliver Jaffe and Olivia Watkins and Olivier Godement and Owen Campbell-Moore and Patrick Chao and Paul McMillan and Pavel Belov and Peng Su and Peter Bak and Peter Bakkum and Peter Deng and Peter Dolan and Peter Hoeschele and Peter Welinder and Phil Tillet and Philip Pronin and Philippe Tillet and Prafulla Dhariwal and Qiming Yuan and Rachel Dias and Rachel Lim and Rahul Arora and Rajan Troll and Randall Lin and Rapha Gontijo Lopes and Raul Puri and Reah Miyara and Reimar Leike and Renaud Gaubert and Reza Zamani and Ricky Wang and Rob Donnelly and Rob Honsby and Rocky Smith and Rohan Sahai and Rohit Ramchandani and Romain Huet and Rory Carmichael and Rowan Zellers and Roy Chen and Ruby Chen and Ruslan Nigmatullin and Ryan Cheu and Saachi Jain and Sam Altman and Sam Schoenholz and Sam Toizer and Samuel Miserendino and Sandhini Agarwal and Sara Culver and Scott Ethersmith and Scott Gray and Sean Grove and Sean Metzger and Shamez Hermani and Shantanu Jain and Shengjia Zhao and Sherwin Wu and Shino Jomoto and Shirong Wu and Shuaiqi and Xia and Sonia Phene and Spencer Papay and Srinivas Narayanan and Steve Coffey and Steve Lee and Stewart Hall and Suchir Balaji and Tal Broda and Tal Stramer and Tao Xu and Tarun Gogineni and Taya Christianson and Ted Sanders and Tejal Patwardhan and Thomas Cunninghman and Thomas Degry and Thomas Dimson and Thomas Raoux and Thomas Shadwell and Tianhao Zheng and Todd Underwood and Todor Markov and Toki Sherbakov and Tom Rubin and Tom Stasi and Tomer Kaftan and Tristan Heywood and Troy Peterson and Tyce Walters and Tyna Eloundou and Valerie Qi and Veit Moeller and Vinnie Monaco and Vishal Kuo and Vlad Fomenko and Wayne Chang and Weiyi Zheng and Wenda Zhou and Wesam Manassra and Will Sheu and Wojciech Zaremba and Yash Patil and Yilei Qian and Yongjik Kim and Youlong Cheng and Yu Zhang and Yuchen He and Yuchen Zhang and Yujia Jin and Yunxing Dai and Yury Malkov}, year={2024}, eprint={2410.21276}, archivePrefix={arXiv}, primaryClass={cs.CL}, url={https://arxiv.org/abs/2410.21276}, }

@article{deepseekai2025deepseekr1incentivizingreasoningcapability, author = {Guo, Daya and Yang, Dejian and Zhang, Haowei and Song, Junxiao and Wang, Peiyi and Zhu, Qihao and Xu, Runxin and Zhang, Ruoyu and Ma, Shirong and Bi, Xiao and Zhang, Xiaokang and Yu, Xingkai and Wu, Yu and Wu, Z F and Gou, Zhibin and Shao, Zhihong and Li, Zhuoshu and Gao, Ziyi and Liu, Aixin and Xue, Bing and Wang, Bingxuan and Wu, Bochao and Feng, Bei and Lu, Chengda and Zhao, Chenggang and Deng, Chengqi and Ruan, Chong and Dai, Damai and Chen, Deli and Ji, Dongjie and Li, Erhang and Lin, Fangyun and Dai, Fucong and Luo, Fuli and Hao, Guangbo and Chen, Guanting and Li, Guowei and Zhang, H and Xu, Hanwei and Ding, Honghui and Gao, Huazuo and Qu, Hui and Li, Hui and Guo, Jianzhong and Li, Jiashi and Chen, Jingchang and Yuan, Jingyang and Tu, Jinhao and Qiu, Junjie and Li, Junlong and Cai, J L and Ni, Jiaqi and Liang, Jian and Chen, Jin and Dong, Kai and Hu, Kai and You, Kaichao and Gao, Kaige and Guan, Kang and Huang, Kexin and Yu, Kuai and Wang, Lean and Zhang, Lecong and Zhao, Liang and Wang, Litong and Zhang, Liyue and Xu, Lei and Xia, Leyi and Zhang, Mingchuan and Zhang, Minghua and Tang, Minghui and Zhou, Mingxu and Li, Meng and Wang, Miaojun and Li, Mingming and Tian, Ning and Huang, Panpan and Zhang, Peng and Wang, Qiancheng and Chen, Qinyu and Du, Qiushi and Ge, Ruiqi and Zhang, Ruisong and Pan, Ruizhe and Wang, Runji and Chen, R J and Jin, R L and Chen, Ruyi and Lu, Shanghao and Zhou, Shangyan and Chen, Shanhuang and Ye, Shengfeng and Wang, Shiyu and Yu, Shuiping and Zhou, Shunfeng and Pan, Shuting and Li, S S and Zhou, Shuang and Wu, Shaoqing and Yun, Tao and Pei, Tian and Sun, Tianyu and Wang, T and Zeng, Wangding and Liu, Wen and Liang, Wenfeng and Gao, Wenjun and Yu, Wenqin and Zhang, Wentao and Xiao, W L and An, Wei and Liu, Xiaodong and Wang, Xiaohan and Chen, Xiaokang and Nie, Xiaotao and Cheng, Xin and Liu, Xin and Xie, Xin and Liu, Xingchao and Yang, Xinyu and Li, Xinyuan and Su, Xuecheng and Lin, Xuheng and Li, X Q and Jin, Xiangyue and Shen, Xiaojin and Chen, Xiaosha and Sun, Xiaowen and Wang, Xiaoxiang and Song, Xinnan and Zhou, Xinyi and Wang, Xianzu and Shan, Xinxia and Li, Y K and Wang, Y Q and Wei, Y X and Zhang, Yang and Xu, Yanhong and Li, Yao and Zhao, Yao and Sun, Yaofeng and Wang, Yaohui and Yu, Yi and Zhang, Yichao and Shi, Yifan and Xiong, Yiliang and He, Ying and Piao, Yishi and Wang, Yisong and Tan, Yixuan and Ma, Yiyang and Liu, Yiyuan and Guo, Yongqiang and Ou, Yuan and Wang, Yuduan and Gong, Yue and Zou, Yuheng and He, Yujia and Xiong, Yunfan and Luo, Yuxiang and You, Yuxiang and Liu, Yuxuan and Zhou, Yuyang and Zhu, Y X and Huang, Yanping and Li, Yaohui and Zheng, Yi and Zhu, Yuchen and Ma, Yunxian and Tang, Ying and Zha, Yukun and Yan, Yuting and Ren, Z Z and Ren, Zehui and Sha, Zhangli and Fu, Zhe and Xu, Zhean and Xie, Zhenda and Zhang, Zhengyan and Hao, Zhewen and Ma, Zhicheng and Yan, Zhigang and Wu, Zhiyu and Gu, Zihui and Zhu, Zijia and Liu, Zijun and Li, Zilin and Xie, Ziwei and Song, Ziyang and Pan, Zizheng and Huang, Zhen and Xu, Zhipeng and Zhang, Zhongyu and Zhang, Zhen}, doi = {10.1038/s41586-025-09422-z}, issn = {1476-4687}, journal = {Nature}, number = {8081}, pages = {633--638}, title = {{DeepSeek-R1 incentivizes reasoning in LLMs through reinforcement learning}}, url = {https://doi.org/10.1038/s41586-025-09422-z}, volume = {645}, year = {2025} }

@misc{openai2025competitiveprogramminglargereasoning, title={Competitive Programming with Large Reasoning Models}, author={OpenAI and : and Ahmed El-Kishky and Alexander Wei and Andre Saraiva and Borys Minaiev and Daniel Selsam and David Dohan and Francis Song and Hunter Lightman and Ignasi Clavera and Jakub Pachocki and Jerry Tworek and Lorenz Kuhn and Lukasz Kaiser and Mark Chen and Max Schwarzer and Mostafa Rohaninejad and Nat McAleese and o3 contributors and Oleg Mürk and Rhythm Garg and Rui Shu and Szymon Sidor and Vineet Kosaraju and Wenda Zhou}, year={2025}, eprint={2502.06807}, archivePrefix={arXiv}, primaryClass={cs.LG}, url={https://arxiv.org/abs/2502.06807}, }

@inproceedings{hu2021loralowrankadaptationlarge, author = {Edward J. Hu and Yelong Shen and Phillip Wallis and Zeyuan Allen{-}Zhu and Yuanzhi Li and Shean Wang and Lu Wang and Weizhu Chen}, title = {LoRA: Low-Rank Adaptation of Large Language Models}, booktitle = {The Tenth International Conference on Learning Representations, {ICLR} 2022, Virtual Event, April 25-29, 2022}, publisher = {OpenReview.net}, year = {2022}, url = {https://openreview.net/forum?id=nZeVKeeFYf9}, bibsource = {dblp computer science bibliography, https://dblp.org} }

@misc{grattafiori2024llama3herdmodels, title={The Llama 3 Herd of Models}, author={Aaron Grattafiori and Abhimanyu Dubey and Abhinav Jauhri and Abhinav Pandey and Abhishek Kadian and Ahmad Al-Dahle and Aiesha Letman and Akhil Mathur and Alan Schelten and Alex Vaughan and Amy Yang and Angela Fan and Anirudh Goyal and Anthony Hartshorn and Aobo Yang and Archi Mitra and Archie Sravankumar and Artem Korenev and Arthur Hinsvark and Arun Rao and Aston Zhang and Aurelien Rodriguez and Austen Gregerson and Ava Spataru and Baptiste Roziere and Bethany Biron and Binh Tang and Bobbie Chern and Charlotte Caucheteux and Chaya Nayak and Chloe Bi and Chris Marra and Chris McConnell and Christian Keller and Christophe Touret and Chunyang Wu and Corinne Wong and Cristian Canton Ferrer and Cyrus Nikolaidis and Damien Allonsius and Daniel Song and Danielle Pintz and Danny Livshits and Danny Wyatt and David Esiobu and Dhruv Choudhary and Dhruv Mahajan and Diego Garcia-Olano and Diego Perino and Dieuwke Hupkes and Egor Lakomkin and Ehab AlBadawy and Elina Lobanova and Emily Dinan and Eric Michael Smith and Filip Radenovic and Francisco Guzmán and Frank Zhang and Gabriel Synnaeve and Gabrielle Lee and Georgia Lewis Anderson and Govind Thattai and Graeme Nail and Gregoire Mialon and Guan Pang and Guillem Cucurell and Hailey Nguyen and Hannah Korevaar and Hu Xu and Hugo Touvron and Iliyan Zarov and Imanol Arrieta Ibarra and Isabel Kloumann and Ishan Misra and Ivan Evtimov and Jack Zhang and Jade Copet and Jaewon Lee and Jan Geffert and Jana Vranes and Jason Park and Jay Mahadeokar and Jeet Shah and Jelmer van der Linde and Jennifer Billock and Jenny Hong and Jenya Lee and Jeremy Fu and Jianfeng Chi and Jianyu Huang and Jiawen Liu and Jie Wang and Jiecao Yu and Joanna Bitton and Joe Spisak and Jongsoo Park and Joseph Rocca and Joshua Johnstun and Joshua Saxe and Junteng Jia and Kalyan Vasuden Alwala and Karthik Prasad and Kartikeya Upasani and Kate Plawiak and Ke Li and Kenneth Heafield and Kevin Stone and Khalid El-Arini and Krithika Iyer and Kshitiz Malik and Kuenley Chiu and Kunal Bhalla and Kushal Lakhotia and Lauren Rantala-Yeary and Laurens van der Maaten and Lawrence Chen and Liang Tan and Liz Jenkins and Louis Martin and Lovish Madaan and Lubo Malo and Lukas Blecher and Lukas Landzaat and Luke de Oliveira and Madeline Muzzi and Mahesh Pasupuleti and Mannat Singh and Manohar Paluri and Marcin Kardas and Maria Tsimpoukelli and Mathew Oldham and Mathieu Rita and Maya Pavlova and Melanie Kambadur and Mike Lewis and Min Si and Mitesh Kumar Singh and Mona Hassan and Naman Goyal and Narjes Torabi and Nikolay Bashlykov and Nikolay Bogoychev and Niladri Chatterji and Ning Zhang and Olivier Duchenne and Onur Çelebi and Patrick Alrassy and Pengchuan Zhang and Pengwei Li and Petar Vasic and Peter Weng and Prajjwal Bhargava and Pratik Dubal and Praveen Krishnan and Punit Singh Koura and Puxin Xu and Qing He and Qingxiao Dong and Ragavan Srinivasan and Raj Ganapathy and Ramon Calderer and Ricardo Silveira Cabral and Robert Stojnic and Roberta Raileanu and Rohan Maheswari and Rohit Girdhar and Rohit Patel and Romain Sauvestre and Ronnie Polidoro and Roshan Sumbaly and Ross Taylor and Ruan Silva and Rui Hou and Rui Wang and Saghar Hosseini and Sahana Chennabasappa and Sanjay Singh and Sean Bell and Seohyun Sonia Kim and Sergey Edunov and Shaoliang Nie and Sharan Narang and Sharath Raparthy and Sheng Shen and Shengye Wan and Shruti Bhosale and Shun Zhang and Simon Vandenhende and Soumya Batra and Spencer Whitman and Sten Sootla and Stephane Collot and Suchin Gururangan and Sydney Borodinsky and Tamar Herman and Tara Fowler and Tarek Sheasha and Thomas Georgiou and Thomas Scialom and Tobias Speckbacher and Todor Mihaylov and Tong Xiao and Ujjwal Karn and Vedanuj Goswami and Vibhor Gupta and Vignesh Ramanathan and Viktor Kerkez and Vincent Gonguet and Virginie Do and Vish Vogeti and Vítor Albiero and Vladan Petrovic and Weiwei Chu and Wenhan Xiong and Wenyin Fu and Whitney Meers and Xavier Martinet and Xiaodong Wang and Xiaofang Wang and Xiaoqing Ellen Tan and Xide Xia and Xinfeng Xie and Xuchao Jia and Xuewei Wang and Yaelle Goldschlag and Yashesh Gaur and Yasmine Babaei and Yi Wen and Yiwen Song and Yuchen Zhang and Yue Li and Yuning Mao and Zacharie Delpierre Coudert and Zheng Yan and Zhengxing Chen and Zoe Papakipos and Aaditya Singh and Aayushi Srivastava and Abha Jain and Adam Kelsey and Adam Shajnfeld and Adithya Gangidi and Adolfo Victoria and Ahuva Goldstand and Ajay Menon and Ajay Sharma and Alex Boesenberg and Alexei Baevski and Allie Feinstein and Amanda Kallet and Amit Sangani and Amos Teo and Anam Yunus and Andrei Lupu and Andres Alvarado and Andrew Caples and Andrew Gu and Andrew Ho and Andrew Poulton and Andrew Ryan and Ankit Ramchandani and Annie Dong and Annie Franco and Anuj Goyal and Aparajita Saraf and Arkabandhu Chowdhury and Ashley Gabriel and Ashwin Bharambe and Assaf Eisenman and Azadeh Yazdan and Beau James and Ben Maurer and Benjamin Leonhardi and Bernie Huang and Beth Loyd and Beto De Paola and Bhargavi Paranjape and Bing Liu and Bo Wu and Boyu Ni and Braden Hancock and Bram Wasti and Brandon Spence and Brani Stojkovic and Brian Gamido and Britt Montalvo and Carl Parker and Carly Burton and Catalina Mejia and Ce Liu and Changhan Wang and Changkyu Kim and Chao Zhou and Chester Hu and Ching-Hsiang Chu and Chris Cai and Chris Tindal and Christoph Feichtenhofer and Cynthia Gao and Damon Civin and Dana Beaty and Daniel Kreymer and Daniel Li and David Adkins and David Xu and Davide Testuggine and Delia David and Devi Parikh and Diana Liskovich and Didem Foss and Dingkang Wang and Duc Le and Dustin Holland and Edward Dowling and Eissa Jamil and Elaine Montgomery and Eleonora Presani and Emily Hahn and Emily Wood and Eric-Tuan Le and Erik Brinkman and Esteban Arcaute and Evan Dunbar and Evan Smothers and Fei Sun and Felix Kreuk and Feng Tian and Filippos Kokkinos and Firat Ozgenel and Francesco Caggioni and Frank Kanayet and Frank Seide and Gabriela Medina Florez and Gabriella Schwarz and Gada Badeer and Georgia Swee and Gil Halpern and Grant Herman and Grigory Sizov and Guangyi and Zhang and Guna Lakshminarayanan and Hakan Inan and Hamid Shojanazeri and Han Zou and Hannah Wang and Hanwen Zha and Haroun Habeeb and Harrison Rudolph and Helen Suk and Henry Aspegren and Hunter Goldman and Hongyuan Zhan and Ibrahim Damlaj and Igor Molybog and Igor Tufanov and Ilias Leontiadis and Irina-Elena Veliche and Itai Gat and Jake Weissman and James Geboski and James Kohli and Janice Lam and Japhet Asher and Jean-Baptiste Gaya and Jeff Marcus and Jeff Tang and Jennifer Chan and Jenny Zhen and Jeremy Reizenstein and Jeremy Teboul and Jessica Zhong and Jian Jin and Jingyi Yang and Joe Cummings and Jon Carvill and Jon Shepard and Jonathan McPhie and Jonathan Torres and Josh Ginsburg and Junjie Wang and Kai Wu and Kam Hou U and Karan Saxena and Kartikay Khandelwal and Katayoun Zand and Kathy Matosich and Kaushik Veeraraghavan and Kelly Michelena and Keqian Li and Kiran Jagadeesh and Kun Huang and Kunal Chawla and Kyle Huang and Lailin Chen and Lakshya Garg and Lavender A and Leandro Silva and Lee Bell and Lei Zhang and Liangpeng Guo and Licheng Yu and Liron Moshkovich and Luca Wehrstedt and Madian Khabsa and Manav Avalani and Manish Bhatt and Martynas Mankus and Matan Hasson and Matthew Lennie and Matthias Reso and Maxim Groshev and Maxim Naumov and Maya Lathi and Meghan Keneally and Miao Liu and Michael L. Seltzer and Michal Valko and Michelle Restrepo and Mihir Patel and Mik Vyatskov and Mikayel Samvelyan and Mike Clark and Mike Macey and Mike Wang and Miquel Jubert Hermoso and Mo Metanat and Mohammad Rastegari and Munish Bansal and Nandhini Santhanam and Natascha Parks and Natasha White and Navyata Bawa and Nayan Singhal and Nick Egebo and Nicolas Usunier and Nikhil Mehta and Nikolay Pavlovich Laptev and Ning Dong and Norman Cheng and Oleg Chernoguz and Olivia Hart and Omkar Salpekar and Ozlem Kalinli and Parkin Kent and Parth Parekh and Paul Saab and Pavan Balaji and Pedro Rittner and Philip Bontrager and Pierre Roux and Piotr Dollar and Polina Zvyagina and Prashant Ratanchandani and Pritish Yuvraj and Qian Liang and Rachad Alao and Rachel Rodriguez and Rafi Ayub and Raghotham Murthy and Raghu Nayani and Rahul Mitra and Rangaprabhu Parthasarathy and Raymond Li and Rebekkah Hogan and Robin Battey and Rocky Wang and Russ Howes and Ruty Rinott and Sachin Mehta and Sachin Siby and Sai Jayesh Bondu and Samyak Datta and Sara Chugh and Sara Hunt and Sargun Dhillon and Sasha Sidorov and Satadru Pan and Saurabh Mahajan and Saurabh Verma and Seiji Yamamoto and Sharadh Ramaswamy and Shaun Lindsay and Shaun Lindsay and Sheng Feng and Shenghao Lin and Shengxin Cindy Zha and Shishir Patil and Shiva Shankar and Shuqiang Zhang and Shuqiang Zhang and Sinong Wang and Sneha Agarwal and Soji Sajuyigbe and Soumith Chintala and Stephanie Max and Stephen Chen and Steve Kehoe and Steve Satterfield and Sudarshan Govindaprasad and Sumit Gupta and Summer Deng and Sungmin Cho and Sunny Virk and Suraj Subramanian and Sy Choudhury and Sydney Goldman and Tal Remez and Tamar Glaser and Tamara Best and Thilo Koehler and Thomas Robinson and Tianhe Li and Tianjun Zhang and Tim Matthews and Timothy Chou and Tzook Shaked and Varun Vontimitta and Victoria Ajayi and Victoria Montanez and Vijai Mohan and Vinay Satish Kumar and Vishal Mangla and Vlad Ionescu and Vlad Poenaru and Vlad Tiberiu Mihailescu and Vladimir Ivanov and Wei Li and Wenchen Wang and Wenwen Jiang and Wes Bouaziz and Will Constable and Xiaocheng Tang and Xiaojian Wu and Xiaolan Wang and Xilun Wu and Xinbo Gao and Yaniv Kleinman and Yanjun Chen and Ye Hu and Ye Jia and Ye Qi and Yenda Li and Yilin Zhang and Ying Zhang and Yossi Adi and Youngjin Nam and Yu and Wang and Yu Zhao and Yuchen Hao and Yundi Qian and Yunlu Li and Yuzi He and Zach Rait and Zachary DeVito and Zef Rosnbrick and Zhaoduo Wen and Zhenyu Yang and Zhiwei Zhao and Zhiyu Ma}, year={2024}, eprint={2407.21783}, archivePrefix={arXiv}, primaryClass={cs.AI}, url={https://arxiv.org/abs/2407.21783}, }

@inproceedings{warner2024smarterbetterfasterlonger, title = "Smarter, Better, Faster, Longer: A Modern Bidirectional Encoder for Fast, Memory Efficient, and Long Context Finetuning and Inference", author = {Warner, Benjamin and Chaffin, Antoine and Clavi{\'e}, Benjamin and Weller, Orion and Hallstr{\"o}m, Oskar and Taghadouini, Said and Gallagher, Alexis and Biswas, Raja and Ladhak, Faisal and Aarsen, Tom and Adams, Griffin Thomas and Howard, Jeremy and Poli, Iacopo}, editor = "Che, Wanxiang and Nabende, Joyce and Shutova, Ekaterina and Pilehvar, Mohammad Taher", booktitle = "Proceedings of the 63rd Annual Meeting of the Association for Computational Linguistics (Volume 1: Long Papers)", month = jul, year = "2025", address = "Vienna, Austria", publisher = "Association for Computational Linguistics", url = "https://aclanthology.org/2025.acl-long.127/", pages = "2526--2547", ISBN = "979-8-89176-251-0" }

@article{moghe-etal-2025-machine, title = "Machine Translation Meta Evaluation through Translation Accuracy Challenge Sets", author = "Moghe, Nikita and Fazla, Arnisa and Amrhein, Chantal and Kocmi, Tom and Steedman, Mark and Birch, Alexandra and Sennrich, Rico and Guillou, Liane", journal = "Computational Linguistics", volume = "51", number = "1", month = mar, year = "2025", address = "Cambridge, MA", publisher = "MIT Press", url = "https://aclanthology.org/2025.cl-1.4/", doi = "10.1162/coli_a_00537", pages = "73--137" }

@inproceedings{parekh-etal-2024-contextual, title = "Contextual Label Projection for Cross-Lingual Structured Prediction", author = "Parekh, Tanmay and Hsu, I-Hung and Huang, Kuan-Hao and Chang, Kai-Wei and Peng, Nanyun", editor = "Duh, Kevin and Gomez, Helena and Bethard, Steven", booktitle = "Proceedings of the 2024 Conference of the North American Chapter of the Association for Computational Linguistics: Human Language Technologies (Volume 1: Long Papers)", month = jun, year = "2024", address = "Mexico City, Mexico", publisher = "Association for Computational Linguistics", url = "https://aclanthology.org/2024.naacl-long.321/", doi = "10.18653/v1/2024.naacl-long.321", pages = "5738--5757" }

@inproceedings{le2024constraineddecodingcrosslinguallabel, author = {Duong Minh Le and Yang Chen and Alan Ritter and Wei Xu}, title = {Constrained Decoding for Cross-lingual Label Projection}, booktitle = {The Twelfth International Conference on Learning Representations, {ICLR} 2024, Vienna, Austria, May 7-11, 2024}, publisher = {OpenReview.net}, year = {2024}, url = {https://openreview.net/forum?id=DayPQKXaQk}, bibsource = {dblp computer science bibliography, https://dblp.org} }

@inproceedings{chen-etal-2023-frustratingly, title = "Frustratingly Easy Label Projection for Cross-lingual Transfer", author = "Chen, Yang and Jiang, Chao and Ritter, Alan and Xu, Wei", editor = "Rogers, Anna and Boyd-Graber, Jordan and Okazaki, Naoaki", booktitle = "Findings of the Association for Computational Linguistics: ACL 2023", month = jul, year = "2023", address = "Toronto, Canada", publisher = "Association for Computational Linguistics", url = "https://aclanthology.org/2023.findings-acl.357/", doi = "10.18653/v1/2023.findings-acl.357", pages = "5775--5796" }

@inproceedings{vamvas-sennrich-2023-towards, title = "Towards Unsupervised Recognition of Token-level Semantic Differences in Related Documents", author = "Vamvas, Jannis and Sennrich, Rico", editor = "Bouamor, Houda and Pino, Juan and Bali, Kalika", booktitle = "Proceedings of the 2023 Conference on Empirical Methods in Natural Language Processing", month = dec, year = "2023", address = "Singapore", publisher = "Association for Computational Linguistics", url = "https://aclanthology.org/2023.emnlp-main.835/", doi = "10.18653/v1/2023.emnlp-main.835", pages = "13543--13552" }

\onecolumn
\appendix

\section{Details on SwissGov-RSD Dataset Collection Process}
\subsection{Domains}
\label{app:admin-domains}

\begin{table}[ht]
\centering
\small
\begin{tabularx}{\linewidth}{@{}l r X@{}}
\toprule
\textbf{Domain} & \textbf{\# Documents} & \textbf{Description} \\
\midrule
admin.ch & 13 & Main page of the Swiss federal government providing general information about its structure, political system, and key officeholders (Federal Council members).  \\
bfe.admin.ch & 86 & Bundesamt für Energie (BFE): Official site of the Swiss Federal Office of Energy, which is the national center of expertise for energy supply and consumption.  \\
bk.admin.ch & 49 & Bundeskanzlei (BK): Website of the Swiss Federal Chancellery, the staff organization of the federal government (Federal Council), providing details on the Federal Chancellor, the Chancellery’s history, its organizational structure, and its various sections and functions.  \\
seco.admin.ch & 20 & Staatssekretariat für Wirtschaft (SECO): The site of SECO (State Secretariat for Economic Affairs), serving as the federal government’s center of expertise for economic and labor market policy.\\
uvek.admin.ch & 21 & Umwelt, Verkehr, Energie und Kommunikation (UVEK): Official site of the Federal Department of the Environment, Transport, Energy and Communications (DETEC), which functions as Switzerland’s ministry for infrastructure and the environment. \\
sbfi.admin.ch & 21 & Staatssekretariat für Bildung, Forschung und Innovation (SBFI): The website of SERI (State Secretariat for Education, Research and Innovation), the federal agency specialized in national and international policies on education, research and innovation. \\
bakom.admin.ch & 14 & Bundesamt für Kommunikation (BAKOM): Official site of the Federal Office of Communications, the Swiss authority responsible for telecommunications, broadcasting (radio and television), and postal services regulation. \\
\midrule
\textbf{Total} & \textbf{224} & \\
\bottomrule
\end{tabularx}
\caption{Overview of crawled (sub-)domains and their document counts in the final SwissGov-RSD dataset.}
\end{table}

\newpage
\subsection{Annotation Guidelines}
\label{app:guidelines}
\begin{figure}[!htbp]
  \centering
  \includegraphics[page=1, width=\textwidth]{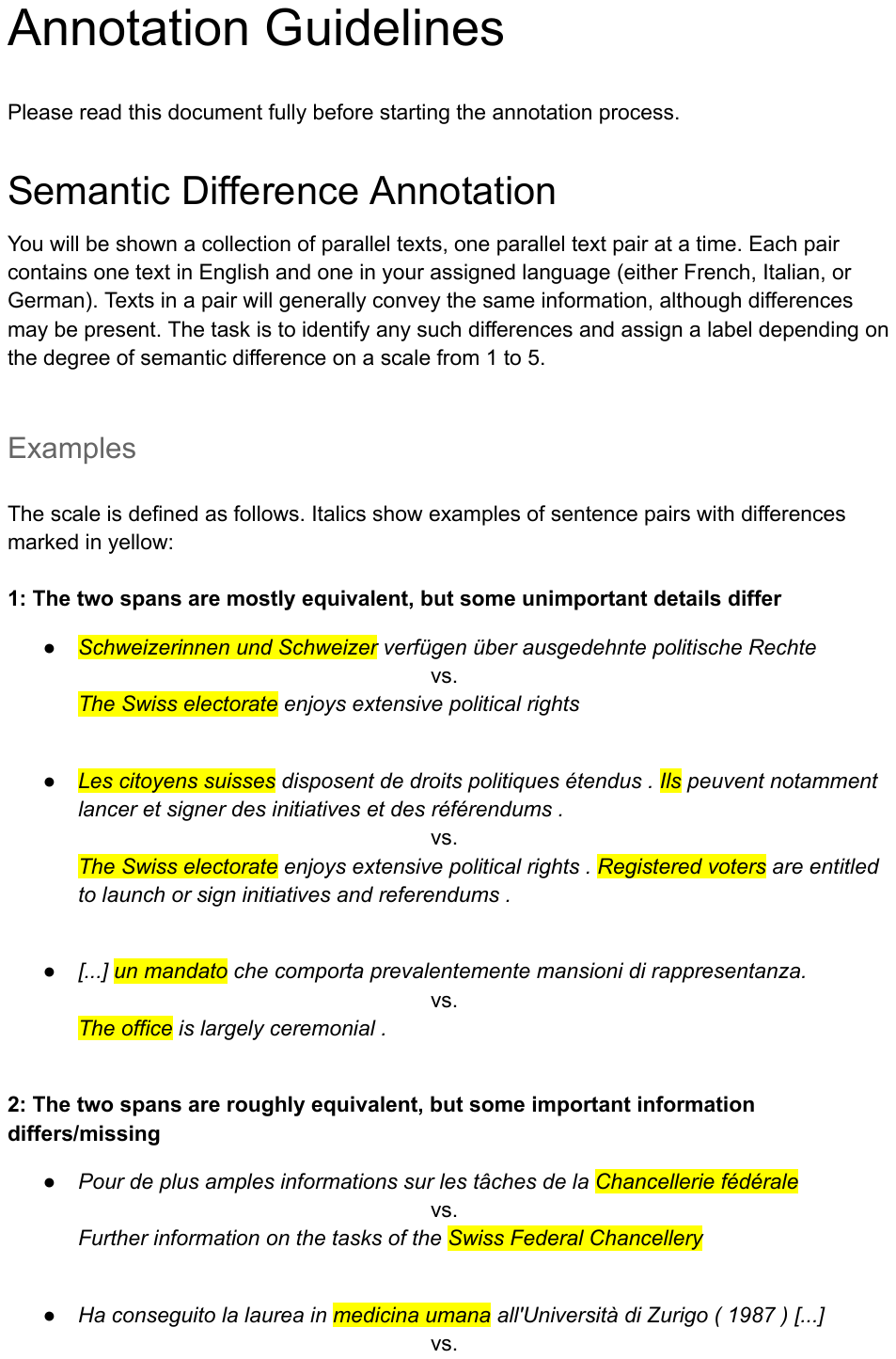}
\end{figure}

\begin{figure}[!htbp]
  \centering
  \includegraphics[page=2, width=\textwidth]{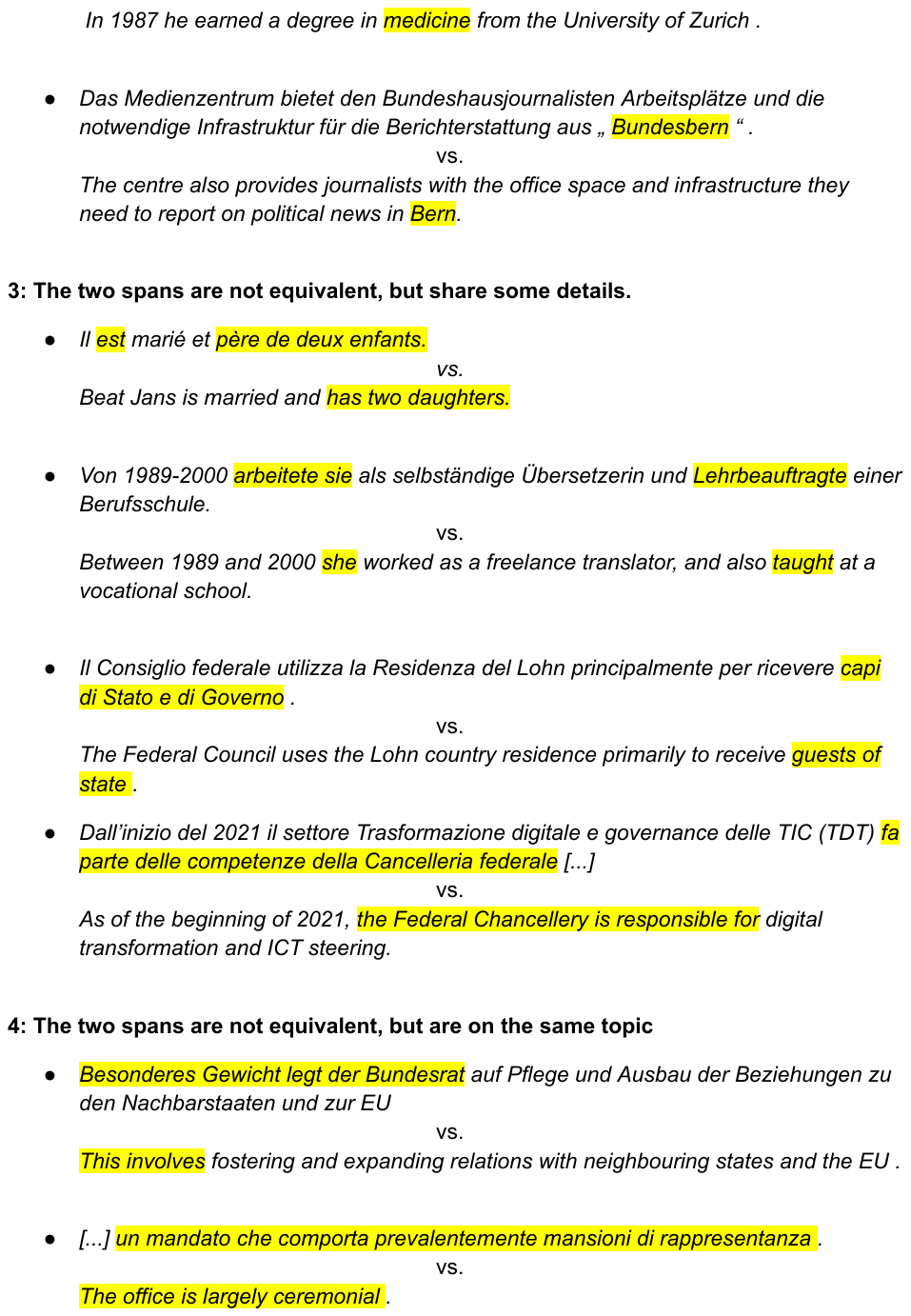}
\end{figure}

\begin{figure}[!htbp]
  \centering
  \includegraphics[page=3, width=\textwidth]{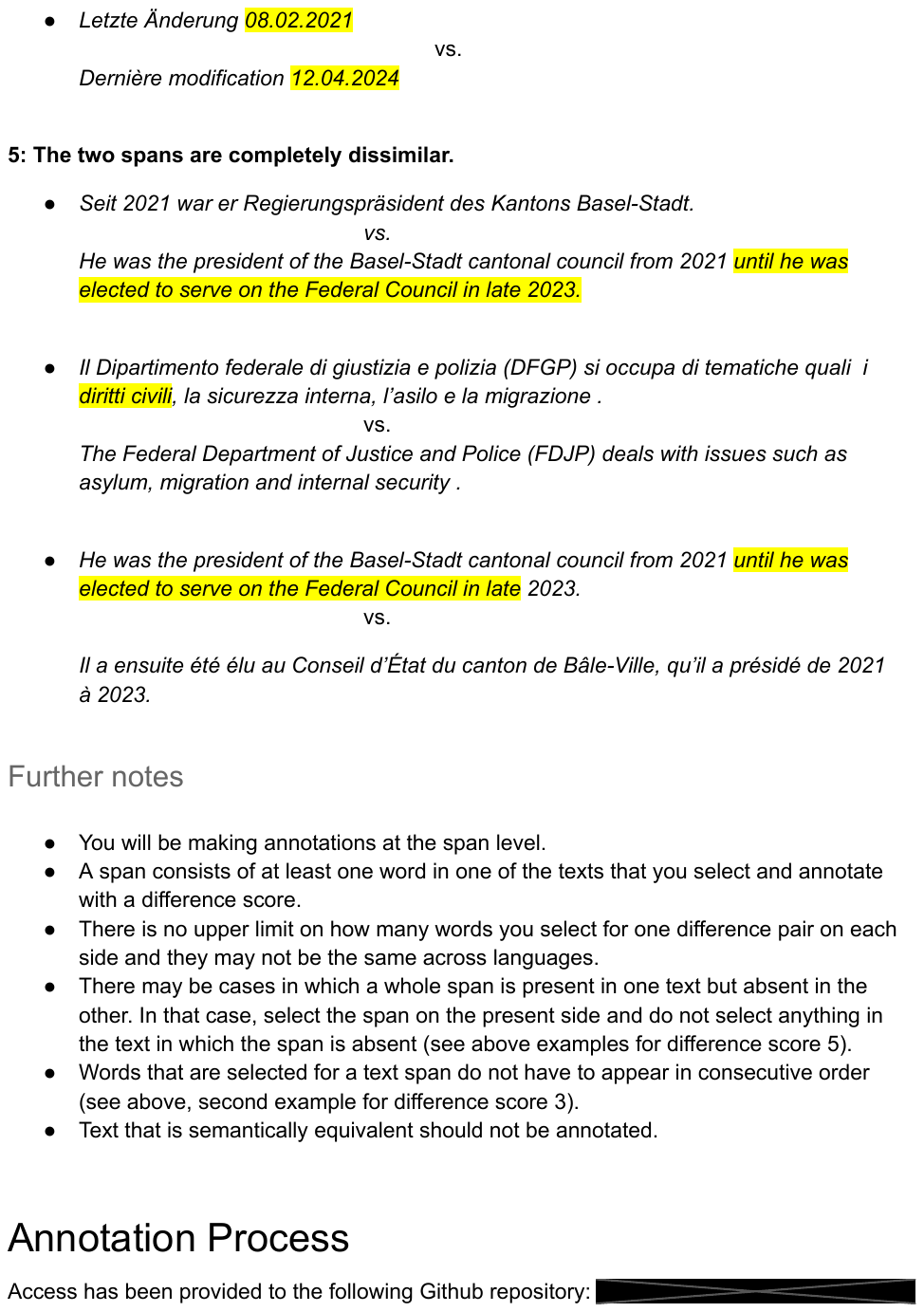}
\end{figure}

\begin{figure}[!htbp]
  \centering
  \includegraphics[page=4, width=\textwidth]{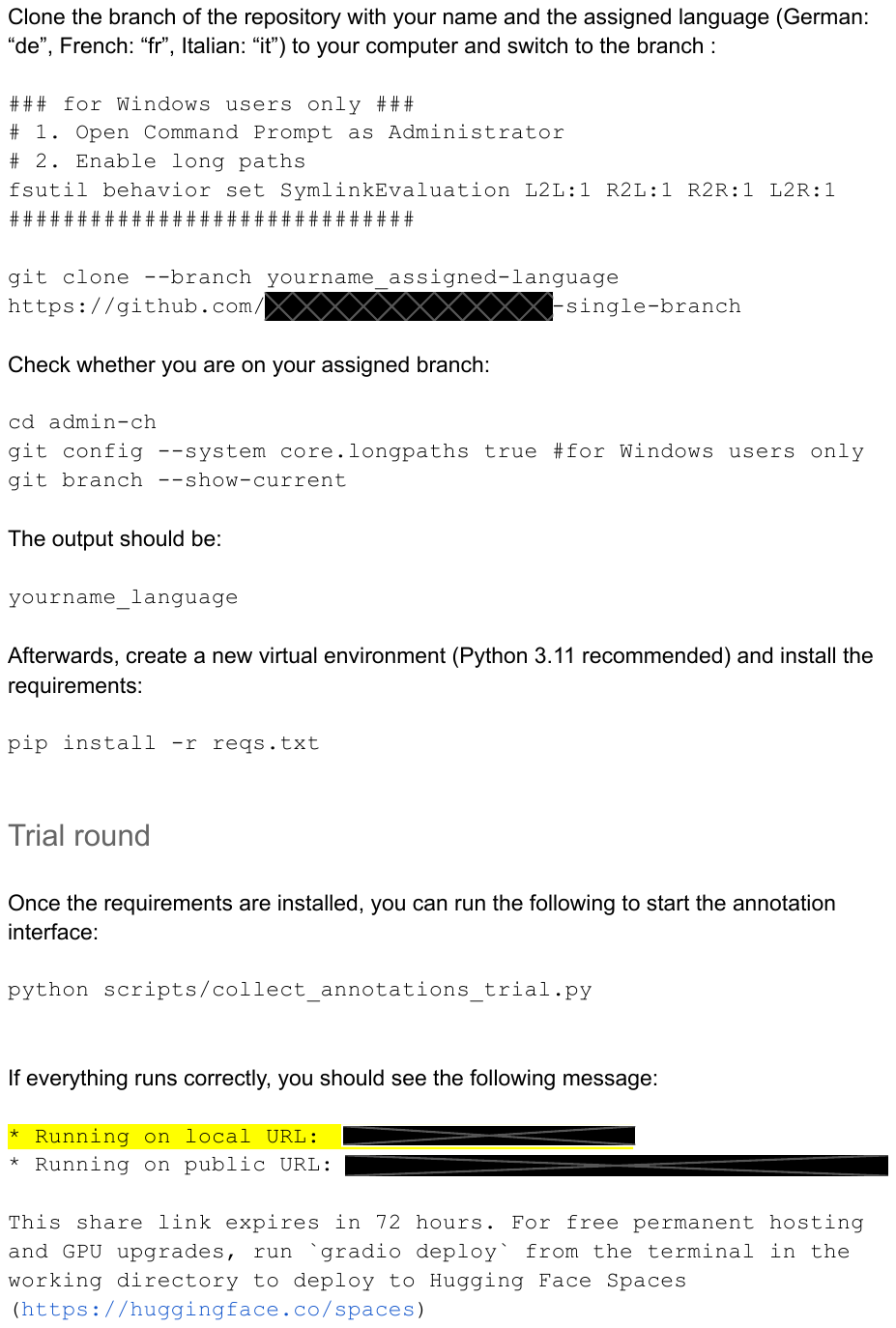}
\end{figure}

\begin{figure}[!htbp]
  \centering
  \includegraphics[page=5, width=\textwidth]{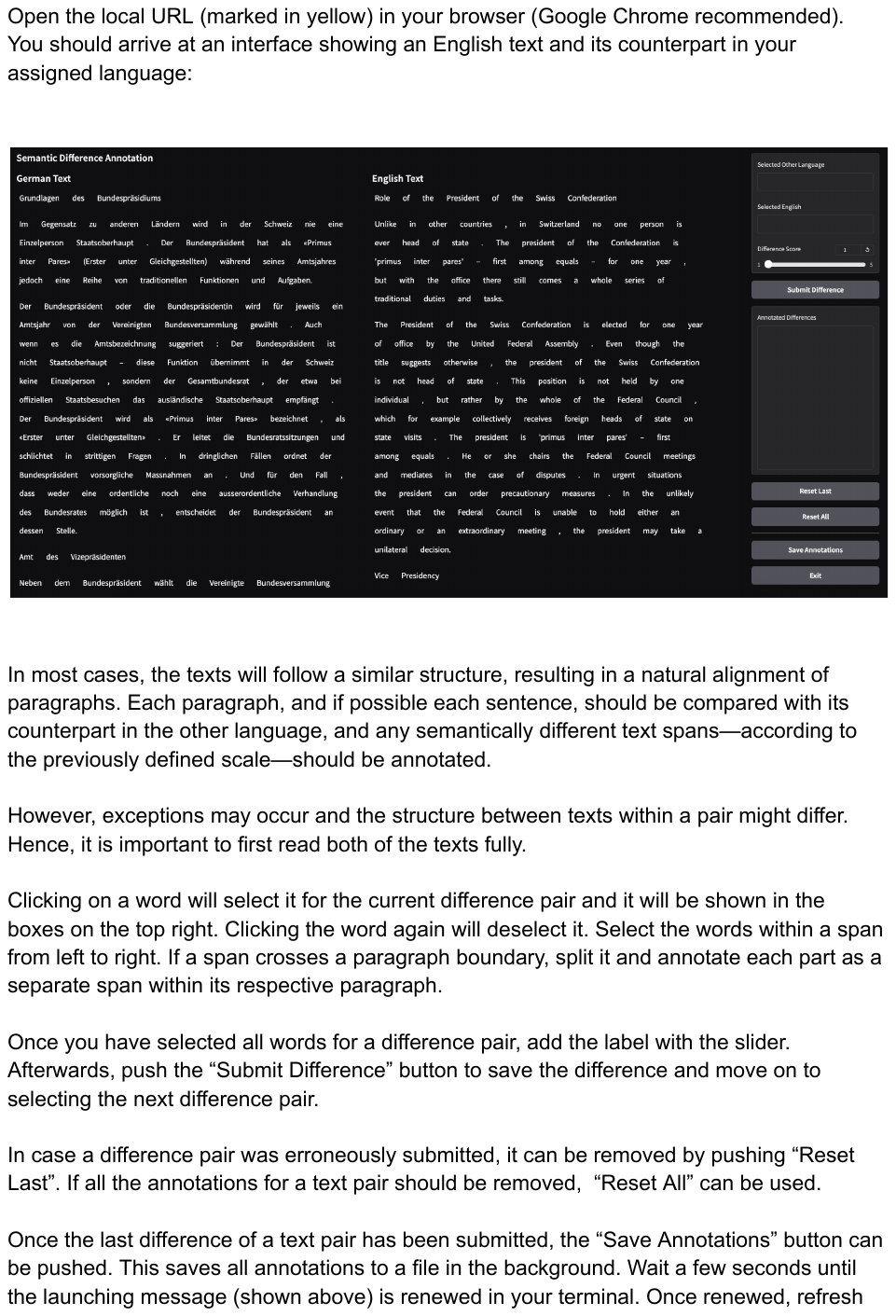}
\end{figure}

\begin{figure}[!htbp]
  \centering
  \includegraphics[page=6, width=\textwidth]{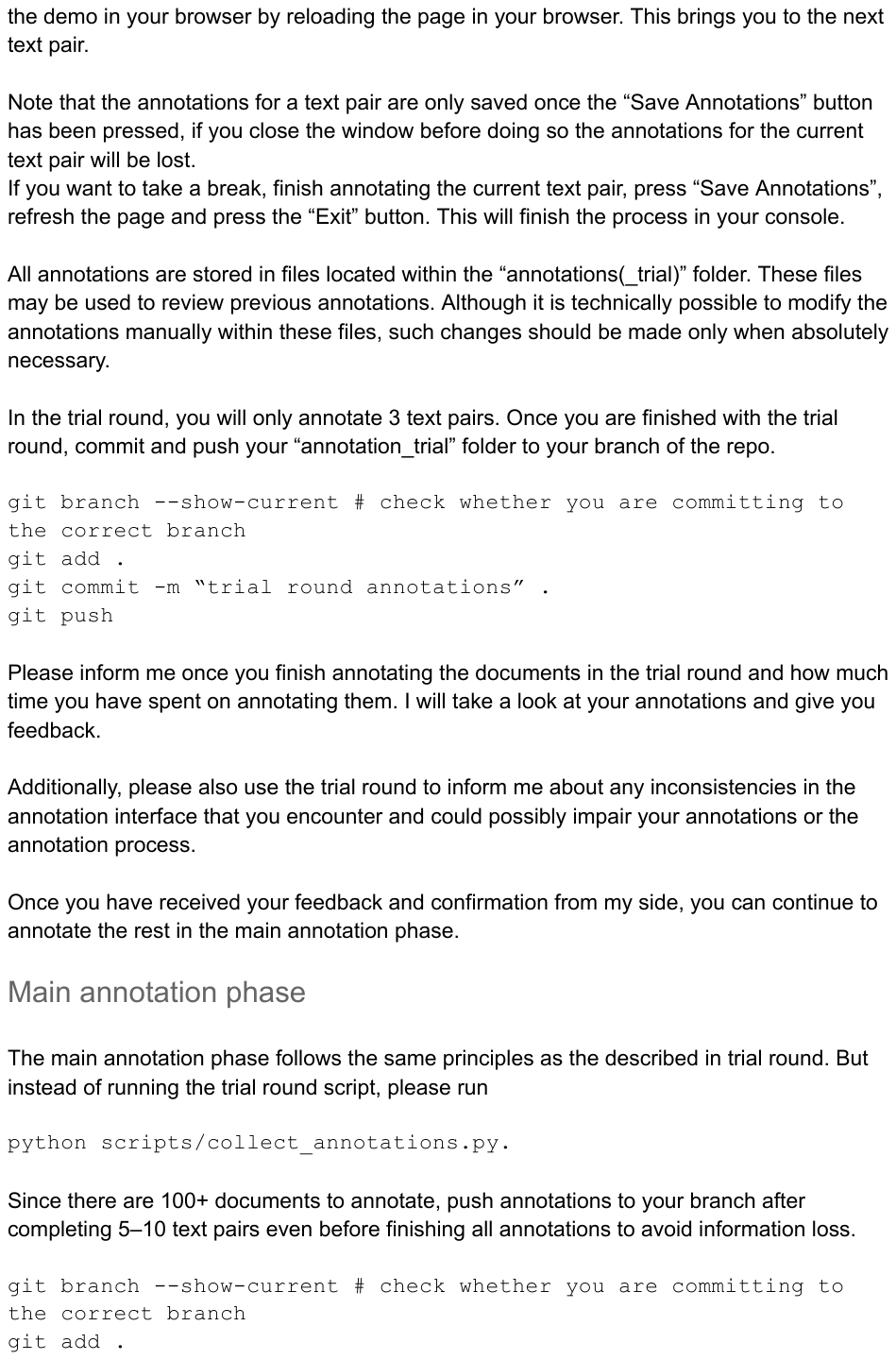}
\end{figure}

\begin{figure}[!htbp]
  \centering
  \includegraphics[page=7, width=\textwidth]{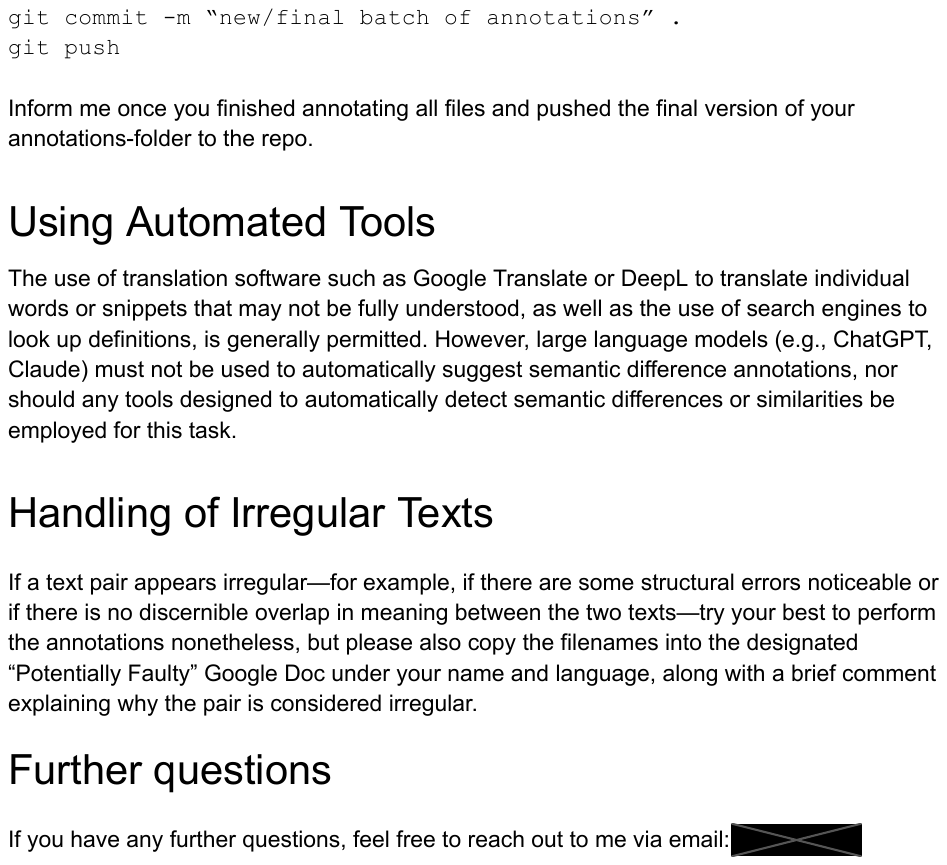}
\end{figure}




\newpage
\subsection{Evaluation}
\label{subsec:ann-eval}


\begin{table}[!htbp]
\centering
\begin{tabular}{lrrr}
\hline
\textbf{Metric} & \textbf{EN-DE} & \textbf{EN-FR} & \textbf{EN-IT} \\
\hline
Total \# differences annotator 1 & 89 & 87 & 99 \\
Total \# differences annotator 2 & 63 & 115 & 105 \\
Total \# differences in tokens annotator 1 & 430 & 1,082 & 772 \\
Total \# differences in tokens annotator 2 & 321 & 821 & 756 \\
\midrule
\# fuzzy matched span pairs >= 50 & 31 & 33 & 43 \\
\# fuzzy matched span pairs >= 75 & 21 & 23 & 34 \\
\# fuzzy matched span pairs >= 90 & 17 & 15 & 30 \\
Exactly matching spans & 16 & 10 & 26 \\
\midrule
Corr. fuzzy matched span pairs >= 50 & 59.97 & 59.12 & 27.43 \\
Corr. fuzzy matched span pairs >= 75 & 78.66 & 63.24 & 31.05 \\
Corr. fuzzy matched span pairs >= 90 & 74.60 & 52.99 & 41.45 \\
Corr. exactly matching spans & 71.22 & 43.00 & 52.64 \\
\midrule
Mean IoU EN & 40.75 & 34.80 & 42.42 \\
Mean IoU OTHER & 32.39 & 34.17 & 40.90 \\
Mean F1 EN & 55.81 & 51.61 & 59.02 \\
Mean F1 OTHER & 46.89 & 49.73 & 56.97 \\
\hline
\end{tabular}
\caption{Results from the automatic evaluation of human annotations in the trial phase.}
\label{tab:trial_round}
\end{table} 

\begin{figure}[!htbp]
  \centering
    \includegraphics[width=0.9\textwidth]{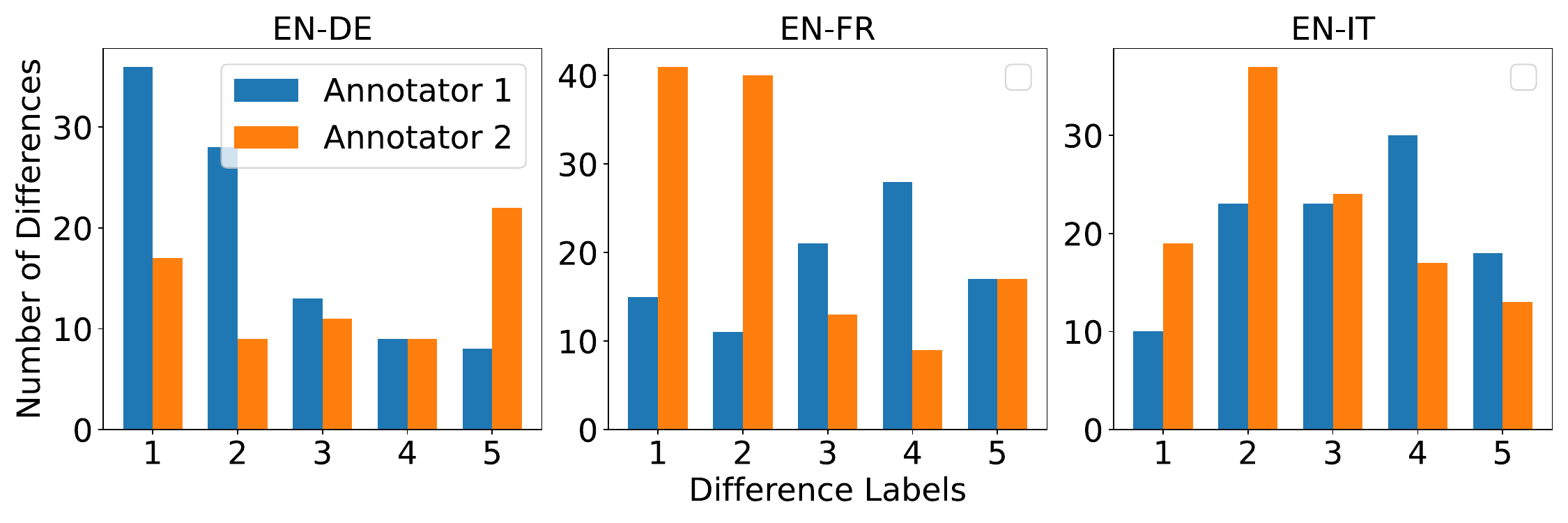}
  \caption{Label distribution for all languages by each annotator in the trial phase.}
\end{figure}

\begin{figure}[!htbp]
  \centering
  \includegraphics[width=0.9\textwidth]{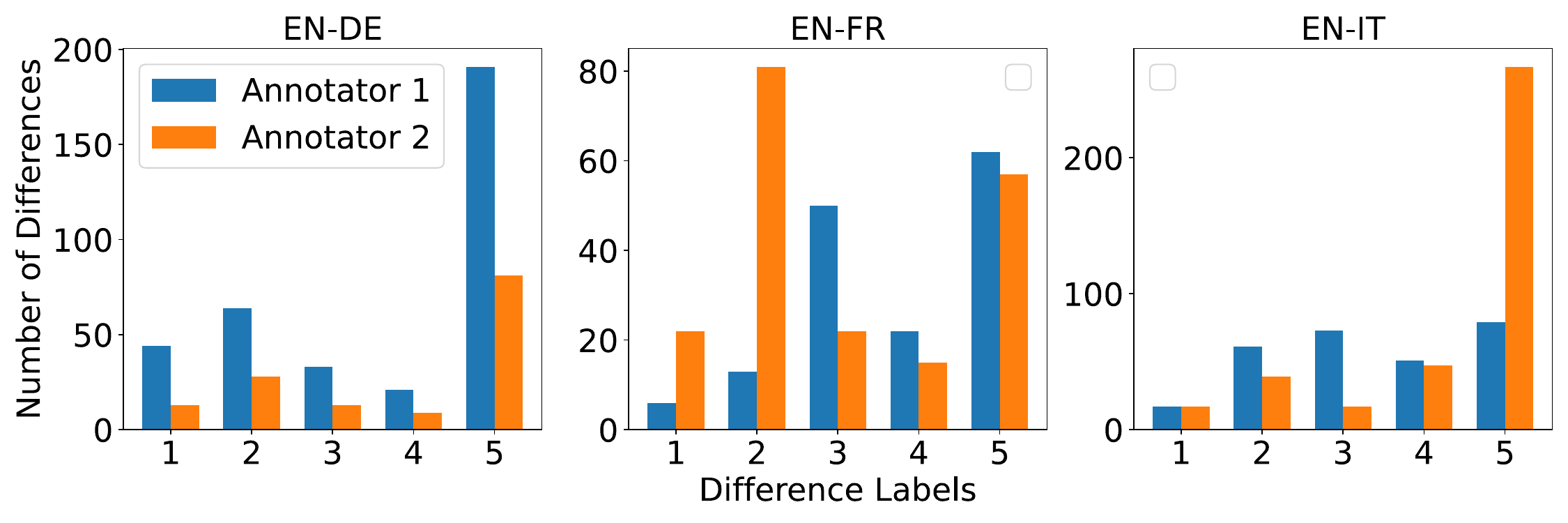}
  \caption{Label distribution for all languages by each annotator in the main phase.}
\end{figure}

\subsection{Impact of Annotator Discrepancies on RSD Evaluation}
\label{app:ann-disc}

\begin{table}[H]
\centering
\small
\begin{tabular}{@{}lrrrrrr@{}}
\toprule
\textbf{Approach} & \multicolumn{2}{c}{\textbf{EN-DE}} & \multicolumn{2}{c}{\textbf{EN-FR}} & \multicolumn{2}{c}{\textbf{EN-IT}} \\
\cmidrule(lr){2-3} \cmidrule(lr){4-5} \cmidrule(lr){6-7}
& \textbf{ann. 1} & \textbf{ann. 2} & \textbf{ann. 1} & \textbf{ann. 2} & \textbf{ann. 1} & \textbf{ann. 2} \\
\midrule
\multicolumn{7}{@{}l@{}}{\textit{DiffAlign (unsupervised)}} \\
DiffAlign XLM-R SimCSE (Spearman) & 0.211 & 0.183 & 0.015 & 0.172 & 0.215 & 0.261 \\
DiffAlign XLM-R SimCSE (Kendall)  & 0.169 & 0.148 & 0.012 & 0.138 & 0.171 & 0.209 \\
\midrule
\multicolumn{7}{@{}l@{}}{\textit{LLMs with few-shot prompting}} \\
GPT-4o (Spearman) & 0.053 & 0.026 & 0.040 & -0.007 & 0.086 & 0.031 \\
GPT-4o (Kendall)  & 0.049 & 0.025 & 0.039 & -0.007 & 0.080 & 0.030 \\
\midrule
\multicolumn{7}{@{}l@{}}{\textit{Fine-tuned encoder models}} \\
ModernBERT (multi) (Spearman) & 0.049 & 0.060 & 0.079 & -0.015 & 0.150 & 0.028 \\
ModernBERT (multi) (Kendall)  & 0.040 & 0.049 & 0.064 & -0.012 & 0.119 & 0.023 \\
\bottomrule
\end{tabular}
\caption{Spearman and Kendall correlations on SwissGov-RSD subsets by the individual annotators. Note that the results for each annotator are for the most part based on different document pairs.}
\label{tab:ann-disc}
\end{table}

The largest absolute difference of the English–French results for the two sets is 0.157 (DiffAlign XLM-R SimCSE (Spearman); 0.015 vs 0.172). We consider both results to be low values, indicating that our main conclusion---systems performing poorly on the document-level RSD task---holds for both annotators.\\

\begin{figure}[!htbp]
  \centering
  \includegraphics[width=\textwidth]{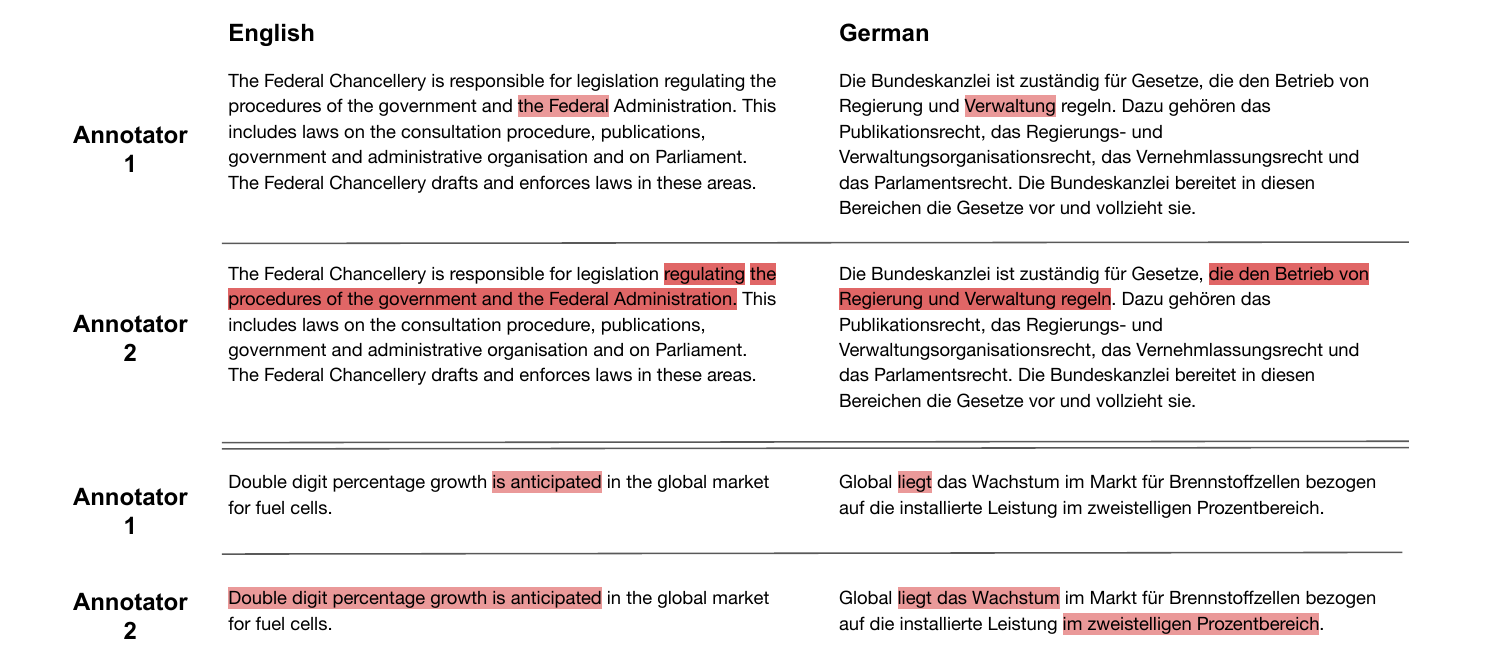}  
  \caption{Different annotation strategies: same difference, but one annotator labels the whole phrase while the other only annotates specific words.}
\label{fig:ann-dis1}
\end{figure}

To make the sources of disagreement between annotators more tangible, in Figure~\ref{fig:ann-dis1} we show excerpts from the overlapping documents with annotations by both annotators. The examples show how differences can arise due to subjective annotation strategies, e.g.: for EN--DE, annotator 1 tends to label semantic differences more minimalistically, while annotator 2 labels the whole phrase in which the difference occurs.\\

\newpage

\section{Description of RSD-iSTS Datasets}
\label{app:description-of-datasets}

\begin{table*}[htb!]
\begin{tabularx}{\textwidth}{@{}Xrrr@{}}
\toprule
Dataset & Human-annotated sentence pairs & Augmentations & Avg. tokens per input \\
\midrule
\mbox{\textit{Training set for encoder models}} \\
Train (EN) & 2800 & 10000 & 67.1 \\
Validation (EN) & 200 & 200 & 62.9 \\
Train (EN--DE*) & 2800 & 10000 & 65.2 \\
Validation (EN--DE*) & 200 & 200 & 67.8 \\
Train (EN--FR*) & 2800 & 10000 & 76.4 \\
Validation (EN--FR*) & 200 & 200 & 80.3 \\
Train (EN--IT*) & 2800 & 10000 & 70.8 \\
Validation (EN--IT*) & 200 & 200 & 74.5 \\
Train (multi*) & 2800 & 30000 & 70.8 \\
Validation (multi*) & 200 & 200 & 74.2 \\
\midrule
\textit{Training set for LLMs} \\
Train (EN) & 2800 & 560 & 67.1 \\
Validation (EN) & 200 & 200 & 62.9 \\
\midrule
\textit{Test set} \\
 iSTS & 100 & 100 & 18.6 \\
\quad + Negatives & 100 & 100 & 30.4 \\
\quad + Documents & 500 & 100 & 156.6 \\
\quad + Permuted & 500 & 100 & 156.6 \\
\quad + Cross-lingual & & & \\
\quad \quad \textsc{en}--\textsc{de} & 500$^\dagger$ & 100 & 154.8 \\
\quad \quad \textsc{en}--\textsc{es} & 500$^\dagger$ & 100 & 164.5 \\
\quad \quad \textsc{en}--\textsc{fr} & 500$^\dagger$ & 100 & 172.6 \\
\quad \quad \textsc{en}--\textsc{ja} & 500$^\dagger$ & 100 & 111.1 \\
\quad \quad \textsc{en}--\textsc{ko} & 500$^\dagger$ & 100 & 119.2 \\
\quad \quad \textsc{en}--\textsc{zh} & 500$^\dagger$ & 100 & 106.0 \\
\quad \quad \textsc{en}--\textsc{it} & 500$^\dagger$ & 100 & 163.7 \\
\bottomrule
\end{tabularx}
\caption{Statistics of the iSTS-RSD-based~\citep{vamvas-sennrich-2023-towards} datasets used for fine-tuning and evaluation. Augmentations are random concatenations of up to 5 human-annotated sentences and do not introduce any new information. $^\dagger$The cross-lingual test sets only have annotations on the English side of the sentence pairs.}
\label{tab:dataset-statistics}
\end{table*}

\section{Example of an Augmented Input}
\label{app:augmented_input}
\begin{figure}[!htb]
\definecolor{berry}{HTML}{B12840}
{\small \textit{\textsf{\colorbox{berry!0}{\strut Couple}\colorbox{berry!0}{\strut sailing}\colorbox{berry!8}{\strut in}\colorbox{berry!8}{\strut a}\colorbox{berry!8}{\strut small}\colorbox{berry!8}{\strut sailboat}\colorbox{berry!0}{\strut .}\colorbox{berry!16}{\strut The}\colorbox{berry!16}{\strut birds}\colorbox{berry!24}{\strut are}\colorbox{berry!24}{\strut swimming}\colorbox{berry!0}{\strut in}\colorbox{berry!0}{\strut the}\colorbox{berry!0}{\strut water}\colorbox{berry!0}{\strut .}\colorbox{berry!8}{\strut Thai}\colorbox{berry!8}{\strut protesters}\colorbox{berry!0}{\strut launch}\\
\colorbox{berry!0}{\strut Bangkok}\colorbox{berry!0}{\strut `}\colorbox{berry!0}{\strut shutdown}\colorbox{berry!0}{\strut '}}}}

\medskip

\noindent{}{\small \textit{\textsf{\colorbox{berry!8}{\strut Thai}\colorbox{berry!8}{\strut opposition}\colorbox{berry!8}{\strut protesters}\colorbox{berry!0}{\strut begin}\colorbox{berry!0}{\strut Bangkok}\colorbox{berry!0}{\strut shutdown}\colorbox{berry!16}{\strut Two}\colorbox{berry!16}{\strut ducks}\colorbox{berry!24}{\strut are}\colorbox{berry!24}{\strut standing}\colorbox{berry!0}{\strut by}\colorbox{berry!0}{\strut the}\colorbox{berry!0}{\strut water}\colorbox{berry!0}{\strut .}\colorbox{berry!0}{\strut Two} \\
\colorbox{berry!0}{\strut people}\colorbox{berry!0}{\strut sailing}\colorbox{berry!8}{\strut a}\colorbox{berry!8}{\strut small}\colorbox{berry!8}{\strut white}\colorbox{berry!8}{\strut sail}\colorbox{berry!8}{\strut boat}\colorbox{berry!0}{\strut .}}}}
\caption{Example of an automatically created augmentation based on three individual, randomly selected sentence pairs. The order of sentences in the two documents has been permuted to make difference recognition more challenging. The highlights visualize the token-level gold labels.}
\label{fig:alignment-example}
\end{figure}

\newpage
\section{English--German Example for Label Projection Prompts}
\label{app:label-projection-prompts}
\small{\texttt{Annotate a pair of input sentences. The goal is to label every word in each sentence regarding its semantic similarity to the words in the other sentence. To help you with that task, you will be given the English translation and the labels for the English translation. Your task is to project the labels of the English translation onto the original sentence.}\\

\noindent\texttt{\#\# Example 1}\\
\texttt{sentence1\_en: "Nevada : 2 dead , 2 hurt in middle school shooting"}\\
\texttt{sentence2\_en: "2 dead , 2 injured in middle school shooting in Nevada"}\\
\texttt{labels1\_en: [5, -1, 5, 5, -1, 5, 5, 5, 5, 5, 5]}\\
\texttt{labels2\_en: [5, 5, -1, 5, 5, 5, 5, 5, 5, 5, 5]}\\

\noindent\texttt{sentence1: "2 dead , 2 injured in middle school shooting in Nevada"}\\
\texttt{sentence2: "Nevada : 2 Tote , 2 Verletzte bei Schießerei an Mittelschule in Nevada"}\\

\noindent\texttt{Output: \{"text\_a": "Nevada : 2 dead , 2 hurt in middle school shooting", "text\_b": "2 Tote , 2 Verletzte bei Schießerei an Mittelschule in Nevada", "labels\_a": [5, -1, 5, 5, -1, 5, 5, 5, 5, 5, 5], "labels\_b": [5, 5, -1, 5, 5, 5, 5, 5, 5, 5, 5]\}}\\

\noindent\texttt{\#\# Example 2}\\
\texttt{sentence1\_en: "sheep standing in a field ."}\\
\texttt{sentence2\_en: "A sheep grazing in a field ."}\\
\texttt{labels1\_en: [4, 2, 5, 5, 5, -1]}\\
\texttt{labels2\_en: [1, 4, 2, 5, 5, 5, -1]}\\

\noindent\texttt{text\_a: "sheep standing in a field ."}\\
\texttt{text\_b: "Ein Schaf grast auf einem Feld ."}\\

\noindent\texttt{Output: \{"text\_a": "Sheep standing in a field .", "text\_b": "Ein Schaf grast auf einem Feld .", "labels\_a": [4, 2, 5, 5, 5, -1], "labels\_b": [1, 4, 2, 5, 5, 5, -1]\}}\\

\noindent\texttt{\#\# Example 3}\\
\texttt{sentence1\_en: "A black dog standing in front of yellow flowers ."}\\
\texttt{sentence2\_en: "A black dog standing in a field ."}\\
\texttt{labels1\_en: [5, 5, 5, 5, 2, 2, 2, 2, 2, -1]}\\
\texttt{labels2\_en: [5, 5, 5, 5, 2, 2, 2, -1]}\\

\noindent\texttt{text\_a: "A black dog standing in front of yellow flowers ."}\\
\texttt{text\_b: "Ein schwarzer Hund steht auf einem Feld ."}\\

\noindent\texttt{Output: \{"text\_a": "A black dog standing in front of yellow flowers .", "text\_b": "Ein schwarzer Hund steht auf einem Feld .", "labels\_a": [5, 5, 5, 5, 2, 2, 2, 2, 2, -1], "labels\_b": [5, 5, 5, 5, 2, 2, 2, -1]\}}\\

\noindent\texttt{\#\# Score scale
\begin{itemize}
  \item \textbf{5:} Complete equivalence
  \item \textbf{3-4:} Very similar or closely similar in terms of semantics.
  \item \textbf{1-2:} Slightly similar or somewhat similar.
  \item \textbf{0:} No relation (there is no word in the other sentence that is even slightly similar in terms of semantics).
  \item \textbf{-1:} Punctuation\\
\end{itemize}}

\noindent\texttt{\#\# Other guidelines
\begin{itemize}
  \item Make sure to output the correct JSON format and to preserve the provided sentence tokenization.\\
\end{itemize}}

\noindent\texttt{
Output only the raw JSON, without any additional text.
}}

\newpage
\section{Few-shot Prompt}
\label{app:few-shot-prompt}
\small{\texttt{Annotate a pair of input sentences. The goal is to label every word in each sentence regarding its semantic similarity to the words in the other sentence.}\\

\noindent\texttt{\#\# Example 1:}\\
\texttt{Input sentence 1: ["Iran", "hopes", "nuclear", "talks", "will", "yield", "\`{}", "roadmap", "'"]}\\
\texttt{Input sentence 2: ["Iran", "Nuclear", "Talks", "in", "Geneva", "Spur", "High", "Hopes"]}\\
\texttt{Output: \{"sentence1": [["Iran", 5], ["hopes", 4], ["nuclear", 5], ["talks", 5], ["will", 3], ["yield", 3], ["\`{}", -1], ["roadmap", 2], ["'", -1]], "sentence2": [["Iran", 5], ["Nuclear", 5], ["Talks", 5], ["in", 0], ["Geneva", 0], ["Spur", 3], ["High", 2], ["Hopes", 2]]\}}\\

\noindent\texttt{\#\# Example 2:}\\
\texttt{Input sentence 1: ["Books", "To", "Help", "Kids", "Talk", "About", "Boston", "Marathon", "News"]}\\
\texttt{Input sentence 2: ["Report", "of", "2", "explosions", "at", "finish", "line", "of", "Boston", "Marathon"]}\\
\texttt{Output: \{"sentence1": [["Books", 1], ["To", 0], ["Help", 0], ["Kids", 0], ["Talk", 0], ["About", 4], ["Boston", 4], ["Marathon", 4], ["News", 4]], "sentence2": [["Report", 1], ["of", 0], ["2", 0], ["explosions", 0], ["at", 0], ["finish", 0], ["line", 0], ["of", 4], ["Boston", 4], ["Marathon", 4]]\}}\\

\noindent\texttt{\#\# Example 3:}\\
\texttt{Input sentence 1: ["Chinese", "shares", "close", "lower", "Wednesday"]}\\
\texttt{Input sentence 2: ["Chinese", "shares", "close", "higher", "Friday"]}\\
\texttt{Output: \{"sentence1": [["Chinese", 5], ["shares", 5], ["close", 5], ["lower", 0], ["Wednesday", 3]], "sentence2": [["Chinese", 5], ["shares", 5], ["close", 5], ["higher", 0], ["Friday", 3]]\}}\\

\noindent\texttt{\#\# Score scale}\\
\texttt{
\begin{itemize}
  \item \textbf{5:} Complete equivalence
  \item \textbf{3-4:} Very similar or closely similar in terms of semantics.
  \item \textbf{1-2:} Slightly similar or somewhat similar.
  \item \textbf{0:} No relation (there is no word in the other sentence that is even slightly similar in terms of semantics).
  \item \textbf{-1:} Punctuation\\
\end{itemize}}

\noindent\texttt{\#\# Other guidelines}\\
\texttt{
\begin{itemize}
  \item Make sure to output the correct JSON format and to preserve the provided sentence tokenization. Output only the raw JSON, without any additional text.
\end{itemize}}

\noindent\texttt{\#\# Input to Annotate}\\
\texttt{Sentence 1: \{\{ sentence1 \}\}}\\
\texttt{Sentence 2: \{\{ sentence2 \}\}}\\

\noindent\texttt{Respond with the JSON object.}
}

\vfill
\pagebreak

\section{Evaluation of projected labels}
\label{app:lp-eval}

\begin{table*}[htb]
\centering
\resizebox{\textwidth}{!}{
\begin{tabular}{@{}lrrrrrr@{}}
\toprule
\textbf{Language Pair} & \textbf{Edited Samples} & \textbf{Total Samples} &
\textbf{Edited Labels} & \textbf{Total Labels} & \textbf{\% Edited Samples} &
\textbf{\% Edited Labels} \\
\midrule
EN--DE & 17 & 50 & 58  & 3379 & 34 & 1.72 \\
EN--IT & 17 & 50 & 81  & 3361 & 34 & 2.41 \\
EN--FR & 23 & 50 & 137 & 3916 & 46 & 3.50 \\
\bottomrule
\end{tabular}
}
\caption{Number of edited labels across language pairs.}
\label{tab:lp-eval}
\end{table*}

\normalsize
\noindent{}We randomly select 50 samples per language pair from the label-projected training data and report the statistics on post-edits that we perform in Table~\ref{tab:lp-eval}. The collected statistics show that while a moderate number of full text samples (34–46\%) contain at least one edit, the proportion of edited labels at the token level remains very low across all language pairs (1.72–3.50\%), suggesting that the LLM-projected labels are largely reliable.

\section{Description of Open-weight Models}
\label{app:description-of-models}

\begin{table*}[htb!]
\begin{tabularx}{\textwidth}{@{}XrrrXr@{}}
\toprule
Name  &
Param. &
Vocab. &
License &
Citation &
  URL \\ \midrule
XLM-R+SimCSE              & 277M  & 250k & MIT & \citet{conneau-etal-2020-unsupervised,vamvas-sennrich-2023-towards} &  \href{https://hf.co/ZurichNLP/unsup-simcse-xlm-roberta-base}{\ExternalLink} \\
ModernBERT-large              & 396M  & 50k & Apache 2.0 & \citet{warner2024smarterbetterfasterlonger} &  \href{https://huggingface.co/answerdotai/ModernBERT-large}{\ExternalLink} \\
EuroBERT 210M             & 210M  & 128k & Apache 2.0 & \citet{Boizard2025-rf} &  \href{https://huggingface.co/EuroBERT/EuroBERT-210m}{\ExternalLink} \\
EuroBERT 610M             & 610M  & 128k & Apache 2.0 & \citet{Boizard2025-rf} &  \href{https://huggingface.co/EuroBERT/EuroBERT-610m}{\ExternalLink} \\
EuroBERT 2.1B             & 2.1B  & 128k & Apache 2.0 & \citet{Boizard2025-rf} &  \href{https://huggingface.co/EuroBERT/EuroBERT-2.1B}{\ExternalLink} \\
MmBERT-small              & 140M  & 256k & MIT & \citet{Marc2025-ea} &  \href{https://huggingface.co/jhu-clsp/mmBERT-small}{\ExternalLink} \\
MmBERT-base              & 307M  & 256k & MIT & \citet{Marc2025-ea} &  \href{https://huggingface.co/jhu-clsp/mmBERT-base}{\ExternalLink} \\
LaBSE              & 500M  & 500k & Apache 2.0 & \citet{feng-etal-2022-language} &  \href{https://huggingface.co/sentence-transformers/LaBSE}{\ExternalLink} \\
bge-m3              & 560M  & 250k & MIT & \citet{chen-etal-2024-m3} &  \href{https://huggingface.co/BAAI/bge-m3}{\ExternalLink} \\
gte-multilingual-base             & 300M  & 129k & Apache 2.0 & \citet{zhang2024mgte} &  \href{https://huggingface.co/Alibaba-NLP/gte-multilingual-base}{\ExternalLink} \\
XLM-R-XL              & 3.48B  & 250k & MIT & \citet{goyal-etal-2021-larger} &  \href{https://huggingface.co/facebook/xlm-roberta-xl}{\ExternalLink} \\
Llama-3.1 8B Instruct              & 8.03B  & 128k & Custom license & \citet{grattafiori2024llama3herdmodels} &  \href{https://huggingface.co/meta-llama/Llama-3.1-8B-Instruct}{\ExternalLink} \\
Llama-3.1 405B Instruct              & 406B  & 128k & Custom license & \citet{grattafiori2024llama3herdmodels} &  \href{https://huggingface.co/meta-llama/Llama-3.1-405B-Instruct}{\ExternalLink} \\
DeepSeek-R1              & 685B  & 129k & MIT & \citet{deepseekai2025deepseekr1incentivizingreasoningcapability} &  \href{https://huggingface.co/deepseek-ai/DeepSeek-R1}{\ExternalLink} \\
\bottomrule
\end{tabularx}
\caption{Number of parameters, vocabulary size and licensing information of the open-weight models used in this paper.}
\label{tab:model-sizes}
\end{table*}

\newpage
\section{Comprehensive iSTS-RSD Result Overview}
\label{app:ists}

\subsection{Results for All Augmentation Categories}

\begin{table*}[ht]
\centering
\small
\resizebox{\textwidth}{!}{%
\begin{tabular}{@{}lrrrrr@{}}
\toprule
\textbf{Approach} & \textbf{iSTS-RSD} & \textbf{+Negatives} & \textbf{+Documents} & \textbf{+Permuted} & \textbf{+Cross-lingual} \\
 & & \small{50\% paraphrases} & \small{5 sentences} & \small{5 inversions} & \small{7 language pairs} \\
 \midrule
\multicolumn{6}{@{}l@{}}{\textit{DiffAlign (unsupervised)}} \\
XLM-R+SimCSE & 64.4 & 62.8 & \underline{56.6} & \underline{54.3} & 36.3 \\
LaBSE 
  & 62.4 & 64.3 & 49.8 & 49.7 & 35.3 \\
bge-m3 
  & \underline{67.4} & \underline{64.4} & 49.5 & 47.9 & \underline{42.5} \\
gte-multilingual-base 
  & 60.1 & 61.1 & 44.8 & 38.5 & 23.6 \\
ModernBERT-large & 52.1 & 47.2 & 49.7 & 48.1 & 12.9 \\
EuroBERT 210M 
  & 38.7 & 47.1 & 52.3 & 51.7 & 27.8 \\
EuroBERT 610M 
  & 30.7 & 38.3 & 41.8 & 41.0 & 24.7 \\
EuroBERT 2.1B 
  & 30.0 & 39.5 & 41.2 & 40.3 & 20.8 \\
mmBERT-small 
  & 54.0 & 49.1 & 46.0 & 46.0 & 17.1 \\
mmBERT-base 
  & 62.3 & 55.4 & 51.3 & 50.5 & 25.5 \\
Qwen3-Embedding 4B 
  & 59.7 & 59.7 & 48.6 & 46.8 & 38.4 \\
Qwen3-Embedding 8B 
  & 56.1 & 53.2 & 48.1 & 45.3 & 37.1 \\
\midrule
\multicolumn{6}{@{}l@{}}{\textit{LLMs with few-shot prompting}} \\
Llama-3.1 8B Instruct & 44.1 & 38.2 & 12.5 & 12.5 & 3.3 \\
Llama-3.1 405B Instruct & 60.6 & 63.3 & 57.8 & 55.2 & 22.8 \\
GPT-4o-mini & 54.6 & 60.5 & 38.0 & 26.2 & 12.4 \\
GPT-4o & \underline{61.1} & \underline{64.9} & \underline{60.7} & \underline{61.2} & 34.8 \\
DeepSeek-R1 & 57.3 & 62.7 & 56.8 & 53.9 & 30.6 \\
o3-mini-low & 59.6 & 64.7 & 58.0 & 58.5 & \underline{36.6} \\
\midrule
\multicolumn{6}{@{}l@{}}{\textit{Fine-tuned LLMs}} \\
Llama-3.1 8B Instruct & 78.0 & 87.6 & 80.9 & 81.0 & 49.2 \\
GPT-4o-mini & \underline{85.9} & \underline{\textbf{92.2}} & \textbf{\underline{88.7}} & \textbf{\underline{87.9}} & \textbf{\underline{62.8}} \\
\midrule
\multicolumn{6}{@{}l@{}}{\textit{Fine-tuned encoder models}} \\
ModernBERT-large (EN) & \textbf{\underline{86.9}} & \underline{84.0} & \underline{81.0} & \underline{81.3} & 51.8 \\
ModernBERT-large (EN-DE*) & 81.5 & 69.4 & 65.0 & 65.2 & 49.3 \\
ModernBERT-large (EN-FR*) & 81.9 & 70.9 & 69.2 & 69.6 & \underline{52.2} \\
ModernBERT-large (EN-IT*) & 82.6 & 72.3 & 68.7 & 69.0 & 48.3 \\
ModernBERT-large (multi*) & 79.2 & 70.0 & 67.2 & 67.4 & 45.4 \\
\bottomrule
\end{tabular}
}
\caption{Comparison of different models and approaches on iSTS-RSD~\citep{vamvas-sennrich-2023-towards}. The table reports token-level Spearman correlations between predicted and gold labels. The variations described in the column headers are cumulative: the rightmost column refers to a cross-lingual test set of permuted documents containing negative examples. The cross-lingual results are averages from the results in Table~\ref{tab:ists-crossling}.
}
\label{tab:ists-results}
\end{table*}
\newpage

\subsection{Results For All iSTS-RSD Language Pairs}
\label{app:cross-ling}

\begin{table}[!htbp]
\centering
\small
\begin{tabular}{@{}lrrrrrrr@{}}
\toprule
\textbf{Approach} & \textbf{EN-DE} & \textbf{EN-FR} & \textbf{EN-IT} & \textbf{EN-ES} & \textbf{EN-JA} & \textbf{EN-KO} & \textbf{EN-ZH} \\
\midrule
\multicolumn{8}{@{}l@{}}{\textit{DiffAlign (unsupervised)}} \\
XLM-R+SimCSE & 44.9 & 45.2 & 44.9 & \underline{47.2} & 15.9 & 33.1 & 23.1 \\
LaBSE
  & 40.6 & 42.3 & 45.3
  & 44.3 & 20.0 & 27.8 & 26.7 \\
bge-m3
  & \underline{47.1} & \underline{45.8} & \underline{47.5}
  & 46.3 & \underline{32.4} & \underline{42.0} & \underline{36.0} \\
gte-multilingual-base
  & 31.2 & 30.2 & 30.5
  & 31.5 & 14.9 & 18.7 & 7.9 \\
ModernBERT-large & 17.3 & 16.7 & 17.2 & 16.4 & 4.2 & 12.4 & 6.1 \\
EuroBERT 210M
  & 32.1 & 34.6 & 33.4
  & 38.1 & 14.7 & 24.0 & 17.4 \\
EuroBERT 610M
  & 29.1 & 30.5 & 31.0
  & 31.4 & 17.1 & 17.2 & 16.9 \\
EuroBERT 2.1B
  & 40.3 & 28.5 & 28.3
  & 30.2 & 11.8 & 9.3 & 12.4 \\
mmBERT-small
  & 23.0 & 23.5 & 22.1
  & 26.7 & 5.8 & 9.6 & 9.3 \\
mmBERT-base
  & 34.0 & 32.6 & 32.6
  & 35.6 & 11.8 & 18.0 & 13.6 \\
Qwen3-Embedding 4B
  & 41.6 & 42.9 & 43.2
  & 42.8 & 29.1 & 37.5 & 31.6 \\
Qwen3-Embedding 8B
  & 40.9 & 40.1 & 41.4
  & 40.8 & 30.3 & 34.2 & 32.2 \\
\midrule
\multicolumn{8}{@{}l@{}}{\textit{LLMs with few-shot prompting}} \\
Llama-3.1 8B Instruct & 2.8 & 3.7 & 2.6 & 7.7 & -2.1 & 7.0 & 1.7 \\
Llama-3.1 405B Instruct & 29.2 & 29.3 & 26.9 & 29.3 & 18.2 & 18.4 & 8.0 \\
GPT-4o-mini & 15.8 & 13.8 & 16.6 & 14.8 & 2.9 & 13.9 & 9.0 \\
GPT-4o & 43.0 & 40.5 & 42.5 & 39.8 & \underline{25.0} & \underline{26.2} & 26.6 \\
DeepSeek-R1 & 38.2 & 40.1 & 34.6 & 39.8 & 19.3 & 19.9 & 22.0 \\
o3-mini-low & \underline{44.8} & \underline{46.5} & \underline{48.2} & \underline{48.5} & 21.9 & 19.2 & \underline{27.1}  \\
\midrule
\multicolumn{8}{@{}l@{}}{\textit{Fine-tuned LLMs}} \\
Llama-3.1 8B Instruct & 66.7 & 67.3 & 66.9 & 68.4 & 19.0 & 29.5 & 26.4 \\
GPT-4o-mini & \textbf{\underline{81.6}} & \textbf{\underline{79.9}} & \textbf{\underline{78.2}} & \textbf{\underline{82.0}} & \underline{35.4} & \underline{39.6} & \underline{41.5} \\
\midrule
\multicolumn{8}{@{}l@{}}{\textit{Fine-tuned encoder models}} \\
ModernBERT-large (EN) & 55.3 & 55.4 & 53.8 & 55.4 & \textbf{\underline{45.9}} & \textbf{\underline{48.8}} & \textbf{\underline{48.2}} \\
ModernBERT-large (EN-DE*) & 58.8 & 58.4 & 58.5 & 58.4 & 32.1 & 43.2 & 37.9 \\
ModernBERT-large (EN-FR*) & 58.4 & 60.7 & 60.1 & \underline{61.2} & 35.8 & 45.9 & 40.7 \\
ModernBERT-large (EN-IT*) & \underline{59.2} & \underline{61.2} & \underline{63.0} & 60.7 & 27.1 & 37.4 & 34.0 \\
ModernBERT-large (multi*) & 54.7 & 57.5 & 58.1 & 57.5 & 25.3 & 35.4 & 30.3 \\
\bottomrule
\end{tabular}
\caption{Token-level Spearman correlations between predicted and gold labels for all individual language pairs covered by the iSTS-RSD. (*) denotes encoders fine-tuned on data with projected labels, \textbf{bold} the best performance overall, and \underline{underline} best performance within model category.}
\label{tab:ists-crossling}
\end{table}

\newpage

\subsection{LLMs Fail to Produce Labels}
\begin{table}[h]
\centering
\begin{tabular}{@{}lr@{}}
\toprule
\textbf{Approach} & \textbf{iSTS-RSD} \\
\midrule
\multicolumn{2}{@{}l@{}}{\textit{LLMs with few-shot prompting}} \\
Llama-3.1 8B Instruct & 10.5\%  \\
Llama-3.1 405B Instruct & 2.0\% \\
GPT-4o & 0.6\% \\
\midrule
\multicolumn{2}{@{}l@{}}{\textit{Fine-tuned LLMs}} \\
Llama-3.1 8B Instruct & 0.6\%\\
GPT-4o-mini & 1.3\% \\
\bottomrule
\end{tabular}
\caption{Percentage of samples for which LLMs failed to produce the correct number of labels for all iSTS-RSD predictions.}
\label{tab:llm-fail-full}
\end{table}

\section{Kendall $\tau$-b Results}
\label{app:kendall-results}
\begin{table*}[h]
\centering
\small
\begin{tabular}{@{}lccc@{}}
\toprule
\textbf{Approach} & \textbf{EN-DE} & \textbf{EN-FR} & \textbf{EN-IT} \\
\midrule
\multicolumn{4}{@{}l@{}}{\textit{DiffAlign (unsupervised)}} \\
DiffAlign XLM-R+SimCSE     & 15.5                & \phantom{0}8.1  & 18.7                \\
DiffAlign ModernBERT       & \phantom{0}1.1      & \phantom{0}0.6  & \phantom{0}2.1      \\
\midrule
\multicolumn{4}{@{}l@{}}{\textit{LLMs with few-shot prompting}} \\
Llama-3.1 405B Instruct    & \phantom{.}-0.8      & \phantom{.}-2.2 & \phantom{0}7.1      \\
GPT-4o                     & \phantom{0}4.0      & \phantom{0}1.2  & \phantom{0}5.1      \\
GPT-4o-mini                & \phantom{0}3.8      & \phantom{0}1.1  & \phantom{0}5.1      \\
\midrule
\multicolumn{4}{@{}l@{}}{\textit{Fine-tuned encoder models}} \\
ModernBERT-large (EN)      & \phantom{0}3.8      & \phantom{0}1.1  & \phantom{0}3.2      \\
ModernBERT-large (EN--DE*)   & \phantom{0}5.2      & \phantom{0}1.3  & \phantom{0}4.5      \\
ModernBERT-large (EN--FR*)   & \phantom{0}4.3      & \phantom{0}4.2  & \phantom{0}7.0                \\
ModernBERT-large (EN--IT*)   & \phantom{0}6.1      & \phantom{0}3.9  & 10.0      \\
ModernBERT-large (multi*)   & \phantom{0}4.6      & \phantom{0}2.0  & \phantom{0}7.1      \\
\bottomrule
\end{tabular}
\caption{Kendall $\tau$-b scores on the SwissGov-RSD task (equivalent to the Spearman correlation in Table~\ref{tab:main-results}). The Kendall scores are generally lower than Spearman scores, but model ranking stays consistent.}
\label{tab:swissgov-results}
\end{table*}

\section{Inference Time Comparison}
\label{app:inference-time}
\begin{table}[!htb]
\begin{tabularx}{\textwidth}{Xrrr}
\toprule
\textbf{Model} & \textbf{Parameters} & \textbf{Short sequence pair (s)} & \textbf{Long sequence pair (s)} \\
\midrule
ModernBERT-large (DiffAlign) & 396M & 0.082 & 0.251 \\
ModernBERT-large (fine-tuned) & 396M & 0.072 & 0.267 \\
Llama-3.1 8B Instr. (few-shot) & 8.03B & 6.400 & 341.900 \\
\bottomrule
\end{tabularx}
\caption{Comparison of inference time when processing either a short sentence pair or a long sequence pair with different approaches.
The table reports average time of 100 inferences for encoder models and 10 inferences for LLMs in seconds measured on a single (encoder) or eight (LLM) NVIDIA GeForce RTX 4090. The short sequence pair has a total token count (separated by whitespaces) of 21, while the long sentence pair counts 962 tokens.}
\label{tab:model-size-time}
\end{table}

\end{document}